\documentclass[preprint,12pt]{elsarticle}

\makeatletter
\def\ps@pprintTitle{%
  \let\@oddhead\@empty
  \let\@evenhead\@empty
  \def\@oddfoot{\reset@font\hfil\thepage\hfil}
  \let\@evenfoot\@oddfoot
}
\makeatother

\usepackage{amssymb}
\usepackage{amsmath}
\usepackage{amsthm}
\usepackage{url}

\usepackage{todonotes}
\usepackage{xcolor}

\definecolor{artemcolor}{HTML}{BB00FF}
\definecolor{andreycolor}{HTML}{00B7BD}

\usepackage{hyperref} 
\usepackage{algorithm}      
\usepackage{algpseudocode}  
\usepackage{amsmath}        

\usepackage{booktabs}
\usepackage{graphicx}
\usepackage{subfig}
\usepackage{bm}

\makeatletter

\newcommand{\fat}[1]{\ifmmode\bm{#1}\else\textbf{#1}\fi}

\newcommand{\set}[1]{\mathbb{#1}}

\newcommand{\setr}[1]{\set{R}^{#1}}

\newcommand{\vect}[1]{\ensuremath{\fat{#1}}}

\newcommand{\matr}[1]{#1}

\newcommand{\func}[1]{#1} 

\newcommand{\loss}[0]{\mathcal{L}}

\newcommand{\vg}[0]{\vect{g}}            
\newcommand{\vu}[0]{\vect{u}}            
\newcommand{\vv}[0]{\vect{v}}            
\newcommand{\vw}[0]{\vect{\omega}}       
\newcommand{\vp}[0]{\vect{\phi}}         
\newcommand{\vx}[0]{\vect{x}}            
\newcommand{\vy}[0]{\vect{y}}            
\newcommand{\vz}[0]{\vect{z}}            

\newcommand{\mA}[0]{\matr{A}}            
\newcommand{\mAw}[0]{\matr{A}_{\vw}}     
\newcommand{\mB}[0]{\matr{B}}            
\newcommand{\mU}[0]{\matr{U}}            
\newcommand{\mS}[0]{\matr{S}}            
\newcommand{\mV}[0]{\matr{V}}            

\newcommand{\ff}[0]{\func{f}}            
\newcommand{\hff}[0]{\widehat{\ff}}      

\newcommand{\fbb}[1][\vw]{\ff_{#1}}

\newcommand{\fsm}[0]{\hff_{\vp}}         

\journal{Neurocomputing}

\newif\ifreviewmode
\newcommand{\rewch}[1]{%
  \ifreviewmode%
    \textcolor{blue}{#1}%
  \else%
    #1%
  \fi%
}
\newcommand{\rewchinp}{%
  \ifreviewmode%
    \color{blue}%
  \fi%
}

\newtheorem{theorem}{Theorem}

\begin{document}

\begin{frontmatter}

\title{Low-rank surrogate modeling and stochastic zero-order optimization for training of neural networks with black-box layers}

\author[airi,sk]{Andrei Chertkov\corref{equal}}
\author[airi,sk]{Artem Basharin\corref{equal}}
\author[sberqtc]{Mikhail Saygin}
\author[airi,sk]{Evgeny Frolov}
\author[sberqtc]{Stanislav Straupe}
\author[airi,sk]{Ivan Oseledets}

\affiliation[airi]{
    organization={Artificial Intelligence Research Institute (AIRI)},
    city={Moscow},
    country={Russia}}
\affiliation[sk]{
    organization={Skolkovo Institute of Science and Technology},
    city={Moscow},
    country={Russia}}
\affiliation[sberqtc]{
    organization={Sber Quantum Technology Center},
    city={Moscow},
    country={Russia}}

\cortext[equal]{ Equal contribution.}

\begin{abstract}
    The growing demand for energy-efficient, high-performance AI systems has led to increased attention on alternative computing platforms (e.g., photonic, neuromorphic) due to their potential to accelerate learning and inference.
    However, integrating such physical components into deep learning pipelines remains challenging, as physical devices often offer limited expressiveness, and their non-differentiable nature renders on-device backpropagation difficult or infeasible.
    This motivates the development of hybrid architectures that combine digital neural networks with reconfigurable physical layers, which effectively behave as black boxes.
    In this work, we present a framework for the end-to-end training of such hybrid networks.
    This framework integrates stochastic zeroth-order optimization for updating the physical layer's internal parameters with a dynamic low-rank surrogate model that enables gradient propagation through the physical layer.
    A key component of our approach is the implicit projector-splitting integrator algorithm, which updates the lightweight surrogate model after each forward pass with minimal hardware queries, thereby avoiding costly full matrix reconstruction.
    We demonstrate our method across diverse deep learning tasks, including: computer vision (deep convolutional network on the CIFAR-10 dataset), audio classification (ECAPA-TDNN model on the UrbanSound8K dataset), and language modeling (a decoder-only transformer model on the FineWeb corpus).
    Notably, across all modalities, the proposed approach achieves near-digital baseline accuracy and consistently enables effective end-to-end training of hybrid models incorporating various non-differentiable physical components (spatial light modulators, microring resonators, and Mach-Zehnder interferometers).
    This work bridges hardware-aware deep learning and gradient-free optimization, thereby offering a practical pathway for integrating non-differentiable physical components into scalable, end-to-end trainable AI systems.
\end{abstract}


\begin{keyword}
    low-rank approximation \sep
    zeroth-order methods \sep
    surrogate modeling \sep
    hardware-aware training \sep
    photonic neural networks.
\end{keyword}

\end{frontmatter}


\section{Introduction}
    \label{sec:intro}

Computational and energy demands of modern deep learning systems are rapidly outpacing the capabilities of conventional digital hardware~\cite{sarker2021deep}.
This has intensified interest in alternative, non-von Neumann computing paradigms capable of providing faster and more energy-efficient machine learning.
Among these, photonic systems offer a particularly promising path toward high-throughput, low-latency, and energy-efficient operation~\cite{bente2025potential}. 

While photonic systems have demonstrated great potential for accelerating inference~\cite{moralis2022neuromorphic,ma2025quantum,ahmed2025universal}, leveraging them during training remains an open challenge~\cite{shastri2021photonics,liao2023integrated,montes2024fundamentals}.
Linear photonic systems~\cite{moralis2024perfect,najjar2024deep}, capable of high-speed matrix-vector multiplication (MVM), are currently more mature and practical than their nonlinear counterparts~\cite{yildirim2024nonlinear,wang2025photonics}, which tend to remain inefficient and challenging to scale.
However, the integration of even these linear components into deep learning pipelines introduces a set of non-trivial problems.
        
Most notably, photonic hardware is inherently non-differentiable, often lacks precise programmability, and can exhibit issues such as noise, drift, or limited control resolution.
These factors render standard backpropagation inapplicable, thereby limiting the ability to train such systems end-to-end.
Moreover, the limited expressiveness of linear photonic models further complicates their direct deployment as plug-in replacements for entirely digital layers.
Taken together, these challenges underscore the critical need for hybrid training pipelines that can effectively combine the raw speed of physical layers with the flexibility of digital deep learning frameworks.
This motivates a central research question for this work:

\begin{figure}[t!]
    \centering
    \includegraphics
        [width=0.95\linewidth]
        {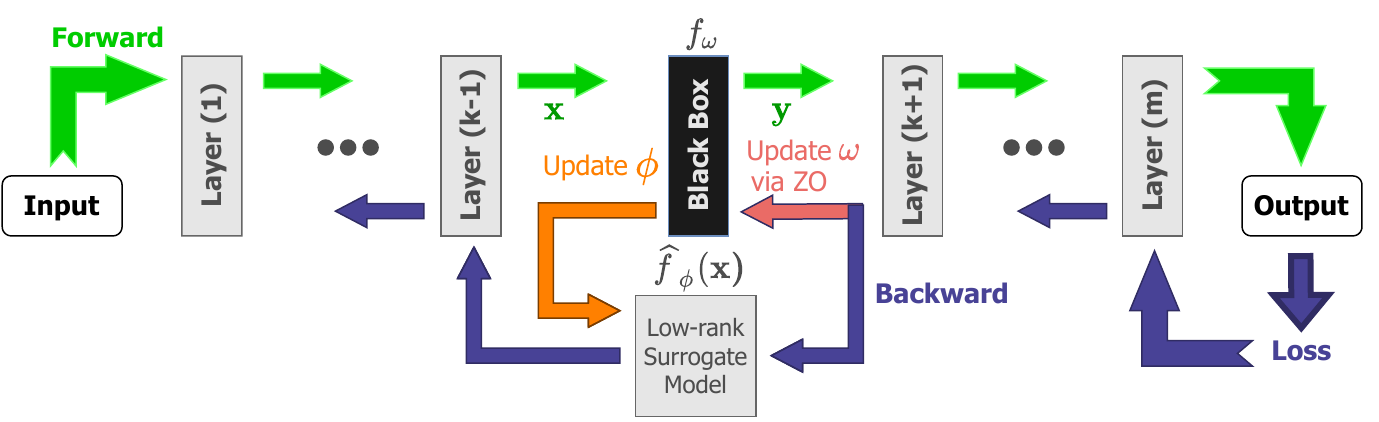}
    \caption{
        \rewch{Schematic representation of the proposed method \emph{astralora}. The diagram distinguishes between the physical BB layer (central block), the digital surrogate optimization loop (bottom path), and the standard digital backpropagation flow (central path).}
    }
    \label{fig:method}
    \vspace{-10pt}
\end{figure}

\begin{center}
    \vspace{20pt}
    \textit{How can we enable efficient, end-to-end training of a hybrid digital-physical neural network when critical components are black boxes, without access to their internal gradients or a pre-calibrated digital twin?}
\end{center}

In this work, we introduce a novel approach \emph{astralora} (Adaptive Surrogate TRAining with LOw RAnk) for training deep neural networks that incorporate a black-box\footnote{
    \rewch{Here, ``black box'' refers to physical components that cannot provide direct gradient information due to their non-differentiable nature or inaccessibility of internal states, making them opaque to standard automatic differentiation.}
} linear photonic layer.\footnote{
    A neural network may contain multiple physical layers, but for simplicity we assume the presence of a single such layer in the subsequent discussion without loss of generality.
}
This layer is treated as a non-differentiable matrix-vector multiplication oracle: $\fbb(\vx) = \mAw \vx$.
Here, the matrix $\mA$ (representing the linear operation) depends on the physical layer's parameters $\vw$ in a complex, non-explicit manner.
To address the challenges of end-to-end training, we propose a two-part solution, schematically represented in Figure~\ref{fig:method}: a dynamically refined low-rank surrogate model $\fsm(\vx) \approx \fbb(\vx)$, which is computationally efficient and enables gradient flow through upstream layers; and a stochastic zeroth-order optimization scheme to update the internal, non-differentiable parameters $\vw$, given that gradients with respect to $\vw$ are unavailable.
Our method employs a parameter-disentangled surrogate $\fsm(\vx)$, which is trained online using a modified (implicit) projector-splitting integrator (PSI) algorithm~\cite{lubich2014projector}.
This surrogate is computationally efficient and readily updated during training, thereby enabling efficient large-scale optimization even with photonic modules incorporated into the system.

We evaluate our approach through extensive experiments, employing deep convolutional models for machine vision and speech recognition tasks. Additionally, we conduct a more complex experiment utilizing a GPT-2–like architecture for a next-token prediction task, where feed-forward blocks are replaced with black-box photonic layers.
Despite relying on approximate surrogate models and zeroth-order updates, our method demonstrates near-parity with fully differentiable baselines across various non-differentiable physical layers (e.g., spatial light modulators, microring resonators, and Mach-Zehnder interferometers).
This crucial finding indicates that accurate backpropagation is not strictly required for successful end-to-end training in such hybrid systems.
The main contributions of our work can be summarized as follows:
\begin{itemize}
    \item
        We propose the implicit projector-splitting integrator (I-PSI) scheme, which enables efficient online retraining of low-rank surrogate models with minimal queries to a non-differentiable (black-box) physical layer.
    \item
        We develop the \emph{astralora} framework, which successfully integrates non-differentiable physical layers into deep learning architectures through a synergistic combination of stochastic zeroth-order optimization and a dynamically updated surrogate model.

    \item
        We conduct comprehensive experimental validation\footnote{
            The code for the \emph{astralora} framework and all numerical examples from this work are publicly available in the repository \url{https://github.com/AndreiChertkov/astralora}.
        } of our approach across three domains: image classification, audio classification, and large-scale language modeling, which demonstrates consistent effectiveness across diverse modalities for multiple types of physical layers.
\end{itemize}

\section{Method}
    \label{sec:method}

In our problem formulation (see Figure~\ref{fig:method}), we consider replacing the $k$-th linear layer of a deep neural network (NN) with a black box (BB) acting as a function:\footnote{
    For brevity, the layer index is omitted in the subsequent formula for the BB action.
}
\begin{equation}
    \vy = \fbb(\vx) = \mAw \vx,
    \quad
    \vx \in \setr{d_{inp}},
    \,\,
    \vy \in \setr{d_{out}},
    \,\,
    \vw \in \setr{d_{bb}},
    \,\,
    \mAw \in \setr{d_{out} \times d_{inp}},
\end{equation}
where $\vx$ is the input vector coming from the previous $(k-1)$th layer, $\vy$ is the output vector coming to the next $(k+1)$th layer, $\vw$ is the set of BB parameters, and $\mAw$ is the matrix corresponding to the BB's linear operation.
Modern deep NNs fundamentally rely on automatic differentiation (autograd) and backpropagation for efficient training.
To effectively leverage gradient-based optimization methods, it is necessary to compute the partial derivative $\partial{\loss}/\partial{w}$ of the loss function $\loss$ with respect to each parameter $w \in \Omega$, where $\Omega$ denotes the complete set of parameters in the NN.
This process, in turn, requires the computation of two types of derivatives for each layer $\ff^{(i)}_{\vw^{(i)}}$ ($i = 1, 2, \ldots, m$) in the NN: $\partial{\ff_{\vw^{(i)}}^{(i)}}/\partial{\vw^{(i)}}$ for updating the layer's parameters, and $\partial{\ff_{\vw^{(i)}}^{(i)}}/\partial{\vx^{(i)}}$ for propagating gradients through the layer.

To update the parameters of the BB (refer to the ``Train'' stage in Figure~\ref{fig:method}), we employ a Monte-Carlo estimation method based on stochastic finite differences, detailed in Section~\ref{ss:train_bb}.
To pass the gradient through the BB during backpropagation (refer to the ``Backward'' stage in Figure~\ref{fig:method}) we utilize a low-rank surrogate model (SM) defined as:
\begin{equation}
    \vy = \fsm(\vx) =
        \mU \mS \mV^T \vx,
    \quad
    \mU \in \setr{d_{out} \times r},
    \,\,
    \mS \in \setr{r \times r},
    \,\,
    \mV \in \setr{d_{inp} \times r}.
\end{equation}
The parameters of the SM, denoted as $\vp = \{ \mU, \mS, \mV \} \in \setr{d_{sm}}$, correspond to a rank-$r$ decomposition of the BB matrix: $\mAw \approx \mU \mS \mV^T$.
Following each training step, the SM parameters $\vp$ are updated using the I-PSI algorithm, as detailed in Section~\ref{ss:update_sm}.
Consequently, in Section~\ref{ss:approach}, we present the \emph{astralora} framework, which integrates non-differentiable physical layers into deep NNs.

\rewch{We note that the choice of a low-rank surrogate model offers several significant advantages for our framework.
Firstly, it provides a differentiable approximation of the BB physical layer, which is essential for enabling gradient flow to upstream digital layers where traditional backpropagation can be applied.
Without this surrogate, end-to-end training of hybrid networks incorporating non-differentiable components would be infeasible. 
Secondly, the low-rank nature ($r \ll \min(d_{inp}, d_{out})$) provides substantial computational and memory efficiency.
The number of parameters in the surrogate ($d_{sm}$) is drastically reduced compared to a full-rank matrix, leading to faster matrix multiplications during forward and backward passes.
This reduced dimensionality also translates to lower memory requirements for storing the surrogate parameters. 
Thirdly, and critically, the low-rank representation is central to the query efficiency of our I-PSI algorithm, detailed in Section 2.2.
This allows us to update the surrogate model with a minimal number of queries to the expensive physical hardware, avoiding costly full matrix reconstructions and making online adaptation practical.
As our experiments demonstrate, even a relatively small rank $r$ can often capture the dominant input-output dynamics of the BB layer sufficiently for effective training, striking an optimal balance between accuracy and resource consumption.}

\subsection{Gradient estimation for the black-box layer} \label{ss:train_bb}

To update the parameters of the BB layer, denoted as $\vw$, we employ stochastic zeroth-order optimization (ZO) approach.
\rewch{This is necessitated by the non-differentiable nature of our BB components, which eliminates the use of standard gradient-based optimization methods.
ZO methods achieve optimization by estimating gradients solely through query-based function evaluations, without requiring explicit derivative computations.
In our framework, this estimation is performed by calculating the action of the gradient operator (i.e., the gradient-vector product), denoted as $\vg(\vx, \vv)$, which acts on an arbitrary error vector $\vv \in \setr{d_{out}}$ originating from the subsequent layer during the backpropagation stage:}
$$
\vg(\vx, \vv) = \left(
    \frac{\partial \fbb(\vx)}{\partial \vw}
\right)^\top \vv.
$$
This expression can alternatively be formulated as the partial derivative of the scalar product:
$$
\vg(\vx, \vv) = \frac{\partial}{\partial \vw}
    \langle \fbb(\vx), \vv \rangle.
$$

Given the unavailability of derivative for $\fbb$ with respect to $\vw$, we approximate the gradient using a stochastic finite-difference method.
Specifically, we employ a well-known formula for the stochastic derivative (e.g., from~\cite{chen2023deepzero}):
$$
\vg(\vx, \vv) \approx \frac{1}{\mu} \mathbb{E}_{\vu} \left(
    \langle \fbb[\vw + \mu \vu](\vx), \, \vv \rangle
    -
    \langle \fbb[\vw](\vx), \vv \rangle
\right) =
    \frac{1}{\mu} \mathbb{E}_{\vu}
        \langle
            \fbb[\vw + \mu \vu](\vx)
            -
            \fbb[\vw](\vx)
            , \,
            \vv
        \rangle,
$$
where $\vu \in \setr{d_{bb}}$ is a random perturbation vector, typically sampled from a standard normal distribution ($\vu \sim \mathcal{N}(0, 1)$), and $\mu > 0$ is a scalar magnitude parameter.
For a practical approximation, the expectation $\mathbb{E}_{\vu}$ is estimated using $M_{BB}$ ($M_{BB} \geq 1$) random samples, leading to the following Monte Carlo approximation for the stochastic derivative:
\begin{equation}\label{eq:stochastic_formula}
\vg(\vx, \vv) \approx
    \frac{1}{\mu \cdot M_{BB}}
    \sum_{i=1}^{M_{BB}}
        \langle
            \fbb[\vw + \mu \vu_i](\vx)
            -
            \fbb[\vw](\vx)
            , \, \vv
        \rangle.
\end{equation}
It is important to note that this formula necessitates $M_{BB} + 1$ queries to the BB for each derivative evaluation.
The minimum number of samples $M_{BB}$ required to ensure sufficient training accuracy for the entire NN, as well as the optimal scalar magnitude $\mu$ for the stability of derivative estimates, are determined empirically.

\subsection{Update of the surrogate model while training} \label{ss:update_sm}

\begin{algorithm}[t!]
\caption{Implicit projector-splitting integrator (I-PSI).}
\label{alg:ipsi}
\begin{algorithmic}[1]

\Require{
    function $\fbb$, that computes BB values;
    rank-$r$ factor matrices $\mU_0$, $\mS_0$, $\mV_0$ of the current SM;
    weights of the BB before update $\vw_0$;
    weights of the BB after update $\vw_1$;
    number of allowed requests to the BB $M_{sm}$
}

\Ensure{
    factor matrices $\mU_1$, $\mS_1$, $\mV_1$ of the updated SM
}


\State{// Compute $
        \matr{P}_1 = \Delta \mA \, \mV_0
    $ by querying the BB \rewch{on columns of $\mV_0$}:}

\For{$j \gets 1 \text{ to } r$}
    \State $
        \vect{v}_j \gets
            \mV_0[:, j]
        $
    
    \State $
        \vect{p}_j \gets
            \fbb[\vw_1](\vect{v}_j)
            -
            \fbb[\vw_0](\vect{v}_j)
    $
\EndFor

\State $
    \matr{P}_1 \gets [
        \vect{p}_1,
        \vect{p}_2,
        \dots,
        \vect{p}_r
]$

\State{// Compute $\mU_1$ \rewch{via QR-based low-rank projector-splitting step}:}

\State $\matr{K}_1 \gets \matr{U}_0\matr{S}_0 + \matr{P_1}$
\Comment{\rewch{intermediate for left singular vectors}}

\State $[\matr{U}_1, \tilde{\matr{S}}_0] \gets \text{qr}(\matr{K}_1)$
\Comment{\rewch{QR yields new orthonormal $\mU_1$}}

\State $\hat{\matr{S}}_0 \gets \tilde{\matr{S}}_0 - \matr{U}_1^\top \matr{P_1}$
\Comment{\rewch{correction for updated singular values}}

\State{// Compute $\matr{P_2} \approx (\Delta \matr{A})^\top \, U_1$ using~\eqref{eq:ipsi_stochastic} and querying the BB:}

\For{$j \gets 1 \text{ to } M_{sm}$}
    \State $\vz_j \gets \mathcal{N}(0, 1)$

    \State $
        \vect{p}_j \gets
            \vz_j
            \left(
            \fbb[\vw_1](\vz_j)
            -
            \fbb[\vw_0](\vz_j)
            \right)^\top
            \mU_1
    $
\EndFor

\State $
    \matr{P}_2 \gets \sum_{j=1}^{M_{sm}}
        \vect{p}_j
$

\State{// Compute $\mV_1$ \rewch{via QR-based low-rank projector-splitting step}:}

\State $\matr{L}_1 \gets \matr{V}_0\hat{\matr{S}}_0^\top + \matr{P_2}$
\Comment{\rewch{intermediate for right singular vectors}}

\State $[\matr{V}_1, \matr{S}_1^\top] \gets \text{qr}(\matr{L}_1)$
\Comment{\rewch{QR yields new orthonormal $\mV_1$}}

\State $\matr{S}_1 \gets (\matr{S}_1^\top)^\top$

\State \Return $\matr{U}_1, \matr{S}_1, \matr{V}_1$

\end{algorithmic}
\end{algorithm}

Following each training step, as the BB parameters evolve to $\vw_{new}$, it becomes necessary to update the low-rank surrogate model (SM) $\hff$, to maintain alignment with $\fbb[\vw_{new}]$.
A naive approach would involve re-initializing the surrogate model, which typically entails computing (either exactly or approximately) the full matrix $\mAw$ from the BB and subsequently performing a Singular Value Decomposition (SVD).
However, this approach is impractical due to significant computational overhead, much of which cannot be efficiently offloaded to optical hardware.
To address this challenge, we propose an extended version of the projector-splitting integrator (PSI)~\cite{lubich2014projector, olaleke2021dynamic}, called the implicit PSI (I-PSI), which is presented in Algorithm~\ref{alg:ipsi}.
This algorithm enables efficient online updating of the low-rank SM with minimal queries to the BB hardware.
The core insight is that, rather than reconstructing the entire matrix $\mAw$, we can efficiently update the existing low-rank decomposition by leveraging the incremental change $\Delta \mA = \mA_{\vw_1} - \mA_{\vw_0}$ between consecutive parameter updates.

The I-PSI algorithm proceeds in four key stages.
First, it computes $\matr{P}_1 = \Delta \mA \, \matr{V}_0$ by evaluating the BB on the  $r$ columns of the current $\matr{V}$ matrix, which requires $r$ queries to the BB.
\rewch{Here, $\matr{P}_1$ specifically captures how this change $\Delta \mA$ affects the output space spanned by the old right singular vectors $\matr{V}_0$, thereby contributing to the update of the left singular vectors.}

Second, it updates the left singular vectors by computing $\matr{U}_1$ through QR decomposition of $\matr{K}_1 = \matr{U}_0 \matr{S}_0 + \matr{P}_1$, followed by a necessary correction to maintain the low-rank structure.
\rewch{The matrix $\matr{K}_1$ is an intermediate structure that combines the previous left singular information ($\matr{U}_0 \matr{S}_0$, which implicitly represents the action of the old surrogate model on the basis $\matr{V}_0$) with the newly computed change $\matr{P}_1$.
The QR decomposition, a fundamental matrix factorization technique, is used here (and analogously for $\matr{V}_1$ later) to compute a new orthonormal basis for the updated low-rank factor from the intermediate, potentially non-orthogonal, matrix $\matr{K}_1$.
This process is crucial for ensuring numerical stability and preserving the manifold structure of rank-$r$ matrices, which is inherent to the projector-splitting integrator.}

Third, we propose to approximate $\matr{P}_2 = (\Delta \mA)^\top \matr{U}_1$ using a stochastic projection method:
\begin{equation}\label{eq:ipsi_stochastic}
(\Delta \mA)^\top \mU \approx
    \sum_{i=1}^{M_{sm}}
        \vz_i (\Delta \mA \vz_i)^\top \mU,
\quad
\vz_i \sim \mathcal{N}(0, 1),
\end{equation}
where $\vz_i \in \setr{d_{inp}}$ is a random normal vector.
\rewch{Similar to $\matr{P}_1$, the matrix $\matr{P}_2$ quantifies the impact of the change $\Delta \mA$, but this time on the transpose of the newly updated left singular vectors $\matr{U}_1$, providing contributions necessary for updating the right singular vectors.}
This step requires $M_{sm}$ random probes to approximate the transpose action without direct access to $(\Delta \mA)^\top$.

Finally, it updates the right singular vectors by computing $\matr{V}_1$ through QR decomposition of $\matr{L}_1 = \matr{V}_0\hat{\matr{S}}_0^\top + \matr{P_2}$.
\rewch{Here, $\matr{L}_1$ is an intermediate matrix for the right singular vectors, constructed by combining the previous right singular information ($\matr{V}_0\hat{\matr{S}}_0^\top$, which incorporates the partially updated singular values $\hat{\matr{S}}_0$) with the change $\matr{P}_2$.
The subsequent QR decomposition then extracts the new orthonormal right singular vectors $\matr{V}_1$ and the final updated singular values $\matr{S}_1$ (derived from $\matr{S}_1^\top$).}

\rewch{
To formally characterize the computational efficiency and the approximation capability of the proposed method, we establish the following theorem regarding the I-PSI update step.}
\begin{theorem}
\label{thm:ipsi_properties}
\rewchinp
Let $\mathcal{M}_r$ be the manifold of matrices of fixed rank $r$. The I-PSI algorithm (Algorithm 1) updates the surrogate model factor matrices from $\mU_0, \mV_0 \in \mathcal{M}_r$ to $\mU_1, \mV_1 \in \mathcal{M}_r$ using exactly
\begin{equation}
N_{queries} = 2r + 2M_{sm}
\end{equation}
forward queries to the black-box physical layer, independent of the matrix dimensions $d_{inp} \times d_{out}$. Furthermore, assuming the stochastic projection in Eq.~\eqref{eq:ipsi_stochastic} is unbiased, the updated surrogate provides a first-order approximation to the projection of the perturbed black-box matrix onto the tangent space of $\mathcal{M}_r$, with an error bound of $O(\|\Delta A\|^2)$, where $\Delta A$ is the change in the physical layer weights.
\end{theorem}

\begin{proof}
\rewchinp
The query complexity follows directly from the construction of the algorithm: step 1 requires $2r$ queries to form $P_1$ (acting on basis $V_0$), and Step 3 requires $2M_{sm}$ queries to approximate the action of $(\Delta A)^\top$ (via random projections).
The approximation property follows from the dynamical low-rank approximation theory established in~\cite{lubich2014projector}, where the splitting integrator is shown to be exact for updates lying within the tangent space, yielding a local error proportional to the curvature of the manifold (second-order in the step size/perturbation $\|\Delta A\|$).
\end{proof}

The proposed approach is particularly well-suited for BB components for several critical reasons.
First, it exclusively requires forward evaluations of the BB, not gradient information, thus making it applicable to physical components such as photonic layers that inherently cannot provide gradient feedback.

Second, it achieves significant query efficiency: \rewch{as follows from Theorem~\ref{thm:ipsi_properties}}, the total query complexity\footnote{
    Algorithm~\ref{alg:ipsi} requires $2r$ queries for matrix-vector products and $2 M_{sm}$ queries for matrix-transpose-vector products.
    For simplicity in discussing the update budget during numerical experiments, we will refer just to $M_{sm}$.
} is $2r+2M_{sm}$, which is substantially lower than the $O(d_{out} \cdot d_{inp})$ queries typically required for full matrix reconstruction (e.g., for full probing or SVD). \rewch{Note that such full scale computations are prohibitive for large layers common in deep NNs (e.g., $512 \times 1024$ or larger). On the other side, low-rank structure drastically reduces this complexity, with small $r \ll \min(d_\text{inp}, d_\text{out})$ empirically sufficient, as NN gradient flow often resides in low-dimensional subspaces~\cite{zhao2024galore}.} 

Third, it enables online adaptation by incrementally updating the existing SM rather than rebuilding it from scratch, thereby efficiently leveraging prior information to maintain accuracy.
It is notable that the theoretical foundations of the projector-splitting approach ensure numerical stability during updates while preserving the manifold structure of rank-$r$ matrices. This property is crucial for maintaining the low-rank constraint throughout the training process.
Empirically, we observe that even with relatively small values of $r$ and moderate query budgets $M_{sm}$, the SM remains sufficiently accurate to enable effective gradient propagation through the NN, as demonstrated by our experimental results in Section~\ref{sec:experiments}.

\subsection{The proposed approach astralora}\label{ss:approach}

\begin{algorithm}[t!]
\caption{The training step with \emph{astralora}.}
\label{alg:astralora}
\begin{algorithmic}[1]

\Require{
    training data batch $(\vx, \vy)$;
    digital parameters $\theta$;
    function $\fbb$, that computes BB values;
    BB parameters $\vw$;
    initial SM factors $\vp=\{ \mU, \mS, \mV \}$;
    learning rates for digital parameters $\eta$, and for BB parameters $\eta_{BB}$;
    perturbation scale $\mu$;
    query budgets $M_{BB}$, and $M_{SM}$;
    loss function $\loss$
}

\Ensure{
    updated $\theta_1$, $\vw_1$, and $\vp_1$
}

\State{// \textbf{Forward pass:}}

\State
    $\vx_{in} \gets \text{NN}_{\text{before}}(\vx)$ 
    \hfill $\triangleright$
        Compute input to BB layer
    
\State
    $\vy_{bb} \gets \fbb(\vx_{in})$
    \hfill $\triangleright$
        Query the physical BB layer

\State
    $\hat{\vy} \gets \text{NN}_{\text{after}}(\vy_{bb})$ 
    \hfill $\triangleright$
        Compute output of the NN
        
\State
    $\mathcal{L} \gets \loss(\hat{\vy}, \vy)$
    \hfill $\triangleright$
        Compute the loss

\State{// \textbf{Backward pass:}}

\State
    $\vv \gets \nabla_{\vy_{bb}} \mathcal{L}$
    \hfill $\triangleright$
        Gradient from downstream layers via autograd

\State
    $\vg_{in} \gets \mV \mS^\top \mU ^\top \vv$
    \hfill $\triangleright$
        Approximate gradient w.r.t. BB input using SM

\State
    $\nabla_{\hat{\theta}} \mathcal{L} \gets \text{Backpropagate} (\vg_{in})$
    \hfill $\triangleright$
        Propagate through upstream layers

\State{// \textbf{Update:}}

\State
    $
        \theta_1 \gets
            \theta
            -
            \eta \cdot \nabla_\theta \mathcal{L}
    $ 
    \hfill $\triangleright$
        Update digital parameters via SGD

\State
    $\vg_{\vw}(\vx, \vv) \approx
        \frac{1}{\mu \cdot M_{BB}}
        \sum_{i=1}^{M_{BB}}
            \langle
                \fbb[\vw + \mu \vu_i](\vx)
                -
                \fbb[\vw](\vx)
                , \, \vv
            \rangle
    $
    \hfill $\triangleright$
        See Section~\ref{ss:train_bb}

\State
    $
        \vw_1 \gets
            \vw
            -
            \eta_{bb} \cdot \textit{mean}_{\vx}
                \left( \vg_{\vw} \right)
    $
    \hfill $\triangleright$
        Zeroth-order update for BB parameters 

\State
    $\vp_1 \gets
        \text{I-PSI}(
            \fbb,
            \vp,
            \vw,
            \vw_1,
            M_{sm}
        )$
    \hfill $\triangleright$
        Update SM (see Algorithm~\ref{alg:ipsi})
        
\end{algorithmic}
\end{algorithm}

Bringing together the steps, described in the previous sections, we summarize the comprehensive training flow in Algorithm~\ref{alg:astralora}.
During the forward pass, the BB layer performs the actual physical computation.
Conversely, the SM is utilized during backpropagation to approximate gradients, as the BB itself is non-differentiable.
The BB parameters are updated via stochastic finite-difference queries.
Subsequently, after each update, the SM is efficiently realigned with the new BB state using the I-PSI algorithm.
Thus, \emph{astralora} seamlessly integrates gradient-free optimization and surrogate modeling into a single, coherent training pipeline, and the central advantage is its query efficiency.
Our approach enables an efficient trade-off between the number of BB queries and the gradient approximation error.
This contrasts with conventional finite-difference-based schemes, which typically require a fixed $O(b \cdot d_{inp})$ queries (where $b$ is the batch size) for error propagation through a black-box layer.

\rewch{While the primary advantage of \emph{astralora} lies in its query efficiency to the non-differentiable BB layer, ensuring the overall computational feasibility of training, especially for large datasets and high-dimensional models, is also crucial. 
Traditional backpropagation, if applicable, would digitally propagate gradients efficiently.
However, for the non-differentiable components that are the focus of our work, backpropagation is fundamentally impossible.
Our method introduces a controlled and scalable computational overhead to specifically handle these components within a training step.}

\rewch{For all digital layers before and after the BB, standard backpropagation incurs the usual computational cost.
For updating the BB parameters, zeroth-order optimization (Section 2.1) requires $M_{BB}$ queries to the physical layer, which is a fixed hyperparameter and does not scale with the dimensionality of the BB parameters $d_{bb}$. 
The surrogate model update, performed by the I-PSI algorithm (Section 2.2 and Algorithm 1), involves $2r + 2M_{sm}$ queries to the physical BB layer.
This query count is significantly lower than the $O(d_{out} \cdot d_{inp})$ queries typically needed for full matrix reconstruction (e.g., via SVD), especially for typical small ranks $r$ and moderate $M_{sm}$.
Furthermore, the digital operations within I-PSI (such as QR decompositions and matrix multiplications) are computationally efficient, as they operate on matrices of reduced dimensions, specifically $d_{out} \times r$ or $d_{inp} \times r$, given that $r \ll \min(d_{inp}, d_{out})$.}

\rewch{In summary, \emph{astralora} introduces a necessary and optimized digital overhead for managing the surrogate model and estimating gradients.
Crucially, the number of expensive queries to the physical BB hardware remains small and controlled, largely independent of the overall model size (e.g., number of parameters of $\omega$) or the dimensions of the layer ($d_{inp}, d_{out}$).
This architecture ensures that the overall training time is primarily dictated by the latency of the physical BB per query, multiplied by a manageable number of queries, rather than by an explosion in digital computation that would render the system impractical.
Our framework thus provides a computationally viable pathway to train hybrid systems where the physical components themselves offer distinct advantages in inference speed or energy efficiency.}

While initially motivated by photonic hardware, the \emph{astralora} scheme is not exclusively restricted to optical systems.
Indeed, any non-differentiable or hardware-constrained module, including neuromorphic accelerators, analogue chips, or other BB hardware can, in principle, be integrated into a trainable NN using the same strategy.
This positions \emph{astralora} as a general framework for hybrid digital–physical learning.
In the subsequent sections, we demonstrate that \emph{astralora} enables end-to-end training across diverse tasks and physical layer implementations, consistently achieving accuracy comparable to fully digital baselines while maintaining query efficiency.

\section{Related work}
    \label{sec:related}

End-to-end training of neural networks with non-differentiable black-box layers presents inherent challenges\rewch{~\cite{ccarpinliouglu2025genetically,momeni2025training}}, primarily because traditional backpropagation cannot propagate gradients through such components.
This section reviews relevant prior work essential to our approach.
We begin by examining gradient-free optimization methods, which, while enabling optimization without explicit gradients, commonly suffer from significant inefficiency at scale.
Next, we discuss low-rank approximation techniques, typically employed to accelerate gradient-based learning in fully differentiable models.
We highlight their inapplicability to true black-box systems, as they rely on an explicitly differentiable matrix.
Finally, we survey existing training approaches for physical neural networks, noting that these often lack efficient and truly joint optimization strategies across their hybrid digital-physical components.
Our proposed \emph{astralora} framework directly addresses these limitations by synergistically combining stochastic zeroth-order optimization with a dynamically updated low-rank surrogate model, thereby facilitating gradient propagation through otherwise non-differentiable components.

\subsection{Gradient-free methods for neural networks}

Backpropagation relies on gradient calculations derived via automatic differentiation, which can encounter challenges when dealing with discontinuous, noisy, or non-differentiable objective functions.
Gradient-free (zeroth-order, or ZO) methods offer viable alternatives, trading computational efficiency for increased robustness to irregular loss surfaces and the ability to optimize non-differentiable components.
Below, we discuss several works that have proposed approaches to improve the performance of ZO methods when applied to modern LLMs.

The work~\cite{liu2018zeroth} pioneered variance-reduced ZO for nonconvex problems, mitigating the high sample complexity of classical ZO methods through accelerated stochastic gradient estimators.
Their key innovation integrates a variance reduction technique with two-point gradient estimators to minimize black-box functions.
While the method is validated effectively on black-box tasks like adversarial attacks and material classification, it remains fundamentally query-inefficient for high-dimensional NN training due to its reliance on iterative forward-pass sampling for gradient approximation, requiring $O(d)$ queries per epoch, where $d$ is the dimensionality.

The authors of~\cite{malladi2023fine} addressed memory constraints with MeZO (memory-efficient zeroth-order optimizer), enabling the fine-tuning of billion-parameter LLMs using only forward passes.
By employing simultaneous perturbation stochastic approximation (SPSA) with in-place perturbation resampling, MeZO achieves a memory footprint equivalent to inference while maintaining performance competitive with backpropagation.
However, MeZO's primary focus on fine-tuning pre-trained models limits its applicability to end-to-end training of networks containing non-differentiable components, as its convergence relies heavily on pre-trained initializations and task prompts.

The work~\cite{chen2024enhancing} proposes LOZO, a memory-efficient ZO optimization method for fine-tuning language models that exploits low-rank gradient structures.
By approximating gradients via low-rank perturbation matrices (LGE) and employing lazy subspace sampling, LOZO captures intrinsic gradient properties while avoiding the need for activation storage.
We note that while LOZO reduces memory overhead for full-model ZO optimization, it lacks mechanisms for localized gradient approximation within non-differentiable components.

The recent work~\cite{chaubard2025scaling} addresses the memory bottleneck in large RNN training by replacing backpropagation through time (BPTT) with Central-Difference Random Gradient Estimation (CD-RGE).
Their method approximates gradients via forward-pass perturbations, eliminating intermediate activation storage and reducing VRAM requirements, thereby enabling billion-parameter RNN training on single GPUs.
While effective for homogeneous RNNs achieving up to 19$\times$ faster convergence than BPTT, their approach requires hundreds of perturbation steps per update, incurring prohibitive computational overhead for systems with slow physical components. 
Furthermore, CD-RGE treats the entire model as a black box, making it difficult to propagate the gradient to upstream layers.

\paragraph{Discussion}
In summary, while recent advances in ZO have improved memory efficiency and scalability for fine-tuning LLMs, critical gaps persist for the end-to-end training of hybrid architectures containing non-differentiable layers.
Variance-reduced methods like~\cite{liu2018zeroth} suffer from prohibitive $O(d)$ query complexity in high dimensions.
Meanwhile, memory-efficient optimizers such as MeZO~\cite{malladi2023fine} and LOZO~\cite{chen2024enhancing} rely on pre-trained initializations and lack mechanisms for localized gradient propagation through individual black-box modules.
Although LOZO exploits low-rank structures for memory reduction, its monolithic gradient estimator cannot backpropagate signals upstream of non-differentiable components.
Similarly, CD-RGE~\cite{chaubard2025scaling} incurs excessive computational overhead for physical systems and treats the entire model as a single black box.
\rewch{Such full-model ZO optimization approaches, while valuable in their specific contexts, would be prohibitively inefficient for our hybrid problem.
This is because they would apply costly query-based gradient estimation to all parameters (including the vast majority in differentiable digital layers), thereby negating the significant efficiency benefits of traditional backpropagation for those differentiable parts of the network.}
Our work bridges these gaps by introducing a dynamic low-rank surrogate framework that enables efficient, layer-wise gradient approximation.
By emulating non-differentiable layers with lightweight linear surrogates after each forward pass, we achieve localized gradient propagation to upstream layers; online retraining capability without pre-training dependencies; and rank-constrained efficiency that avoids $O(d)$ sampling.
This approach uniquely supports the end-to-end optimization of systems integrating physical components while scaling to hundreds of millions of parameters, which is a key advantage over existing ZO methods.

\subsection{Low-rank approximation for backpropagation}

Low-rank approximation techniques have emerged as powerful tools to address computational and memory bottlenecks in backpropagation-based training of deep NNs.
By exploiting intrinsic low-dimensional structures in weight matrices or gradient updates, these methods enable the efficient training of large-scale models while maintaining performance.
This subsection reviews key approaches leveraging low-rank representations to optimize gradient-based learning, focusing on their reliance on end-to-end differentiability, which is a fundamental constraint that limits their applicability in systems containing non-differentiable components.

Authors of~\cite{wang2024lora} propose LoRA-GA, an enhanced initialization scheme for Low-Rank Adaptation (LoRA) that accelerates convergence through gradient alignment (GA). 
The core innovation involves initializing low-rank adapter matrices $U$ and $V$ using SVD of the full-model gradient matrix ($\nabla_A\mathcal{L}$).
This ensures that the first optimization step of the low-rank product $UV$ closely approximates the direction of full fine-tuning gradients, i.e., $\Delta(UV) \approx \zeta\Delta A$, where $A$ is a matrix of weights.
Empirical validation shows LoRA-GA achieves 2-4$\times$ faster convergence than vanilla LoRA.
However, the approach fundamentally relies on explicit gradient computation through all layers during initialization (via SVD of $\nabla_A\mathcal{L}$), making it incompatible with architectures containing non-differentiable black-box components. 

Similarly, the recent work~\cite{zhao2024galore} proposes Gradient Low-Rank Projection (GaLore), a memory-efficient training strategy for LLMs. 
GaLore leverages the intrinsic low-rank structure of gradients during optimization to reduce the memory consumption of optimizer.
Unlike LoRA, which constrains weight updates to a static low-rank subspace, GaLore performs full-parameter learning through dynamic low-rank projections of gradients.
While GaLore offers substantial improvements in memory efficiency for several LLMs, it remains constrained by its reliance on end-to-end differentiability.

The work~\cite{gooneratne2020low} addresses memory constraints in on-device training through low-rank gradient approximation.
By reparameterizing weight gradients as $\nabla\mathcal{L}(A) \approx U\nabla\mathcal{L}(V)^{\top} + \nabla\mathcal{L}(U)V^{\top}$ (where $U$ and $V$ are low-rank factors), the authors effectively leverage low-rank structure in gradient space rather than weight space, preserving model capacity while enabling memory-intensive optimizers on resource-limited devices.
We note that this method fundamentally relies on differentiable operations and backpropagation-compatible components.
This constitutes a significant limitation for our target scenario involving non-differentiable black-box modules (e.g., physical components) where gradients become inaccessible.

The study in~\cite{fournier2023can} proposes using forward gradients with locally supervised auxiliary losses to approximate backpropagation, replacing random directional derivatives with local network gradients to reduce variance.
While effective for standard architectures (e.g., reducing ResNet-18's accuracy gap from $\sim$40\% to $\sim$3\% on CIFAR-10), their method fundamentally assumes differentiability throughout the computational graph and requires architectural modifications to inject auxiliary networks at intermediate layers.

Recent work~\cite{refael2024adarankgrad} introduces AdaRankGrad, a method that exploits the empirical observation that gradient matrices in LLM training exhibit progressively lower rank.
Their approach adaptively projects gradients onto low-rank subspaces during Adam optimization, reducing memory usage by 25-50\% while maintaining performance.
We note that while effective for differentiable networks, AdaRankGrad fundamentally relies on backpropagation-compatible layers.
However, the empirical observation regarding the low rank of gradients in this work is certainly very interesting in the context of our method.
\rewch{These low-rank techniques have also been extended to unsupervised learning settings, such as interpretable neural networks for data clustering via differentiable reconstruction of orthogonal nonnegative matrix factorization (ONMF) and sparse autoencoders~\cite{gai2025interpretable}, and dual-space topological isomorphism for maximizing predictive diversity in unsupervised domain adaptation~\cite{wang2025dual}.}

Follow-up work~\cite{refael2025sumo} proposes SUMO, a subspace-aware optimizer that leverages exact SVD-based orthogonalization of first-order moments within dynamically adapted low-rank subspaces to accelerate memory-efficient LLM training.
By explicitly aligning optimization steps with spectral loss landscape characteristics, SUMO mitigates approximation errors from iterative methods like Newton-Schulz orthogonalization.
It theoretically bounds these errors and demonstrates empirically faster convergence with 20\% memory reduction compared to other methods (e.g., GaLore) in both pre-training and fine-tuning scenarios.
We note that SUMO assumes gradient-based moment updates, which prevents direct application of this approach to NNs containing black-box components.

\paragraph{Discussion}
While the reviewed methods demonstrate compelling advantages in memory efficiency and convergence acceleration for differentiable NNs, they fundamentally depend on end-to-end gradient propagation.
This renders them inapplicable when black-box components (e.g., physical layers) disrupt the computational graph.
Our approach overcomes this limitation through two synergistic innovations.
First, we replace explicit gradient calculations at non-differentiable layers with a lightweight, dynamically updated low-rank surrogate model that enables continuous gradient approximation to upstream layers.
Second, we decouple black-box parameter optimization using stochastic zeroth-order methods, eliminating differentiability requirements downstream.
Crucially, our framework maintains the computational benefits of low-rank representations (similar to methods like GaLore and LoRA-GA) but extends them to hardware-aware scenarios where physical components introduce non-differentiable operations.
The dual strategy of surrogate modeling and gradient-free optimization establishes a new paradigm for end-to-end trainable systems that remains tractable even at scale.
This addresses a critical gap not resolved by existing low-rank approximation techniques.

\subsection{Training approaches for physical neural networks}

Photonic NNs in particular, and analog NNs in general are typically trained by one of three complementary routes, supported by a growing software stack.
First, many systems rely on calibrated digital surrogates and chip-in-the-loop updates: a ``digital twin'' is fitted offline and then mapped to hardware, with measurements used to compensate for mismatch.
A representative example is~\cite{OnChipDiffractive}, which trains with a well-calibrated surrogate and then deploys weights to the chip.
For scalable on-chip adaptation,~\cite{L2ight} propose state calibration, analytical zeroth-order core mapping, and subspace learning with multi-level sparsity to maintain efficient in-situ updates.
When true gradients are difficult to access, zeroth-order training can still be effective at scale; e.g.,~\cite{TensorCompressedTraining} show variance-reduced, tensor-compressed ZO updates that are useful for hardware-aware optimization.
Ideas from the edge computing community, while electronic, can be adopted conceptually for photonics.
E.g.,~\cite{pMetaOnDevice} constrain updates to a subset of parameters for memory and latency efficiency during continual adaptation.

Second, in-situ gradient methods enable the training of photonic NNs without resorting to the calculation of derivatives in the digital electronic domain.
This is achieved by propagating the learning signal through the same optical circuits themselves.
The first experimental demonstration of in-situ backpropagation on a reconfigurable silicon-photonics mesh is presented in~\cite{insitu_backprop_integrated}, which exploits bidirectional propagation and internal taps to measure gradients and perform updates largely on-chip.
For diffractive free-space systems, the theoretical framework for optical adjoints was proposed in~\cite{DNN_backprop_proposal}. 
An alternative that keeps both passes in optics is the end-to-end optical backpropagation, analyzed by~\cite{LvovskySLM}, which leverages saturable-absorber nonlinearities and pump–probe dynamics.
A closely related milestone is a single-chip photonic deep NN trained with forward-only updates~\cite{onchip_forwardonly_englund}.

\begin{sloppypar} 
Third, backpropagation-free, forward-only training algorithms avoid reverse-mode physics entirely and can be robust to nonidealities.
Direct Feedback Alignment replaces symmetric weight transport with fixed random feedback and has been demonstrated in the hybrid electro-photonic training of modern architectures~\cite{dfa_optical_transformer}.
The Forward–Forward algorithm~\cite{ffa_hinton} optimizes per-layer ``goodness'' over positive and negative examples; building on this algorithmic proposal, an optical realization is reported in~\cite{FFA_optical}.
For hardware where gradients are inaccessible or excessively noisy, Multiplexed Gradient Descent (MGD) offers fast ZO updates via multiplexed perturbations~\cite{multiplexed_gd_proposal}.

These methods are enabled by an expanding software ecosystem for differentiable photonics and cross-layer co-design: Neurophox~\cite{pai2019parallel} for unitary/orthogonal meshes and ONN simulation; Photontorch~\cite{laporte2019highly} for PyTorch-native photonic circuit simulation with autograd; PyTorch-ONN~\cite{jiaqigu2021L2ight} for coherent/incoherent ONN layers at scale; SimPhony (ScopeX-ASU)~\cite{SimPhony} for device-circuit-architecture co-modeling; and DAT MPNN~\cite{zheng2023dual} as a reference implementation of Dual Adaptive Training.
\end{sloppypar}

\paragraph{Discussion}
While photonic and analog NNs benefit from multiple complementary training paradigms, each exhibits limitations that constrain end-to-end optimization.
Digital-surrogate methods reliably map software-trained models to hardware but require careful calibration and incur overhead for in-situ adaptation, thereby limiting scalability.
In-situ gradient approaches allow direct gradient propagation on optical circuits, yet they depend on specific architectures and on-chip measurement infrastructure, making them less flexible.
Backpropagation-free, forward-only algorithms offer robustness under inaccessible or noisy gradients, but their convergence and generalization performance can be architecture-dependent and often lag behind gradient-based methods for complex tasks. 
While the growing software ecosystem facilitates simulation, co-design, and hybrid training, no existing approach fully addresses scalability, hardware variability, and training efficiency simultaneously.
Our work aims to bridge these gaps by introducing flexible, hardware-aware surrogate methods that enable localized gradient propagation, efficient online adaptation, and integration of non-differentiable photonic components in large-scale architectures.

\section{Simulated physical layers}
    \label{sec:layers}

This section characterizes the physical photonic devices that constitute the black-box (BB) layers within our hybrid modeling framework.
We investigate five distinct architectures: generic programmable matrix-vector multiplier (MVM); spatial light modulator (SLM) for dynamic wavefront shaping; microring resonator (MRR) banks for wavelength-selective operations; structured matrix layer based on Monarch matrices and SLM blocks, and Mach-Zehnder interferometers (MZIs).

Each layer is characterized by a transfer matrix $\mAw$ that maps an input vector $\vx$ to an output vector, expressed as $\fbb(\vx) = \mAw \vx$.
The relationship between $\mAw$ and the tunable parameters $\vw$ varies across different models, reflecting the diversity of optical hardware implementations.
Consequently, our training approach is evaluated with layers whose transfer matrices range from straightforward to highly intricate functions of $\vw$.
A more detailed description of each layer under consideration\footnote{
    For all considered layers, an additional scalar scale parameter is included (trained gradiently), as the elements of the matrix $\mAw$ have limited values for some layers.
} is provided below.

\subsection{Matvec layer}

A particularly straightforward linear optical layer is one whose transfer-matrix elements can be programmed independently and directly. 
In this case, the trainable weights $\vw$ are mapped onto the matrix
\begin{equation}
    \mAw^{(matvec)} =
        \text{reshape}(\vw, (d_{out}, d_{inp})),
\end{equation}
as in cross-bar–programmable photonic circuits \cite{CoherentXbar}.
We refer to this configuration as the ``matvec'' layer because, while it behaves like a conventional digital matrix–vector multiplication, it is still trained purely as a BB.

\subsection{MRR layer}

\begin{figure}[t!]
    \centering
    \includegraphics[width=1.0\linewidth]{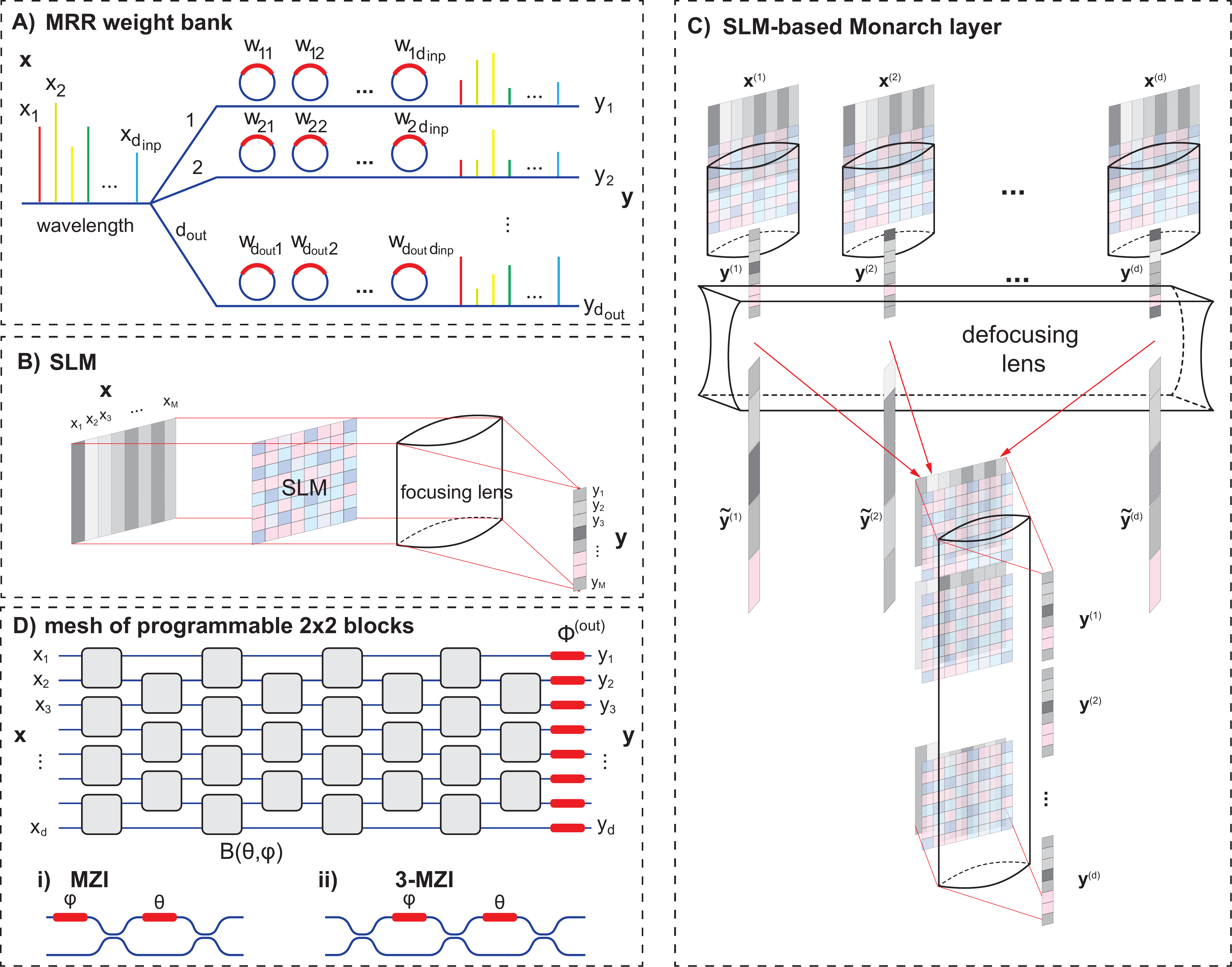}
    \captionsetup{skip=14pt}
    \caption{Illustration of the black-box physical layers used in the hybrid network setup of the \emph{astralora} framework: a) MRR weight banks, b) SLM-based multiplier, c) Monarch matrix multiplier exploiting SLM-based optical blocks, d) planar interferometric meshes of the MZI (i) and $3$-MZI (ii) blocks. \rewch{Each black-box layer implements a non-differentiable linear transformation $\fbb(\vx) = \mAw \vx$, where $\vw$ are the tunable hardware parameters. The linear digital layer in the target neural network can be replaced with one of these layers.}}
    \label{fig:physical_layers}
\end{figure}

In integrated photonics, using microring resonators (MRRs) provides a convenient way of manipulating optical fields in a compact and energy-efficient settings.
Therefore, optical multipliers exploiting programmable MRRs, such as MRR weight banks are considered as one of the architecture of linear layers for NNs~\cite{MRR_weightbanks}.
The MRR linear layer, which we used in our simulations, is shown in Figure~\ref{fig:physical_layers}A and operates as follows.
First, the input vector $\vx = (x_1, x_2, \ldots, x_{d_{inp}})$  is encoded onto a set of optical carriers with centre wavelengths $\lambda_j$, so that the optical field amplitude at wavelength $\lambda_j$ equals the corresponding vector element $x_j$.
These spectral components are then fanned out into $d_\text{out}$ parallel waveguides, each evanescently coupled to a separate row of MRRs.

The $k$-th row contains $d_\text{inp}$ resonators.
Each resonator multiplies one component $x_j$ by its programmed weight $\mA_{kj}$. The value of $\mA_{kj}$ is set by tuning the phase-shift value $w_{kj}$ of the corresponding MRR in the bank, such that $\mA_{kj}=\text{func}(w_{kj})$, where function looks like
\begin{equation}\label{eq:mrr_func}
    \text{func}(w)=2\,\sqrt{\frac{a^2 + r^2 - 2 a r \cos w}{1 + (a r)^2 - 2 a r \cos w}}-1.
\end{equation}
Here, $a$ and $r$ are the intrinsic amplitude transmission and self-coupling coefficients of the MRR, which depend on the physical implementation of the layer. In our experiments, we used representative values $a = 0.8$ and $r = 0.9$.

After modulation, the weighted optical fields in each row are coherently summed, yielding the desired operation.
Overall, the layer implements the linear transformation
\begin{equation}
    \mAw^{(MRR)} =
        \text{reshape}(
            \text{func}(\vw), (d_{out}, d_{inp})).
\end{equation}


\subsection{SLM layer}

Free-space optics provides access to a high-dimensional parameter space, making it well-suited for massively parallel information processing.
It enables linear operations using optical components such as lenses and spatial modulators.
In this work, we consider two optical layers implemented via free-space optics.

The simplest free-space MVM scheme, illustrated in Figure~\ref{fig:physical_layers}B, was first proposed in~\cite{TamuraSLM} and later implemented for reconfigurable large-scale algebraic operations in~\cite{LvovskySLM}.
This optical arrangement utilizes a spatial light modulator (SLM) combined with a cylindrical lens.
The SLM comprises a 2D array of programmable pixels capable of modulating the phase profile of an incoming optical beam. 
The phase shift values represent the trainable weights of the linear layer.

To multiply an input real-valued vector $\vx$ by a $d_{out} \times d_{inp}$ matrix $\mAw^{(SLM)}$, the input vector is encoded as $d_{inp}$ optical field amplitudes with corresponding amplitude values $x_i$ ($i = 1, 2, \ldots, d_{inp}$).
Each amplitude is spatially expanded vertically to overlap with corresponding columns of SLM pixels, each column having a size of $d_{out}$.
A cylindrical lens then focuses the optical field horizontally onto a narrow vertical slit.
In this dimension, the lens performs a Fourier transform, and the slit selects only the zero spatial frequency component.
Immediately after the slit, the complex optical fields are described by:
\begin{equation}
    z_j =
        \frac{1}{\sqrt{d_{inp}}}
        \sum_{i=1}^{d_{inp}}
            e^{i\theta_{ji}} x_i,
\end{equation}
where $\theta_{ji} = \text{reshape}(\vw,(d_{out}, d_{inp}))$ represents the real phase shifts applied by the SLM. 
Assuming measurement of the real components of the optical fields $z_j$, we define the output as $y_j = \text{Re}(z_j)$ for $j = 1, 2, \ldots, d_{out}$.
Thus, the transfer matrix of the SLM layer is given by:
 \begin{equation}
     \mAw^{(SLM)} =
        \frac{1}{\sqrt{d_{inp}}}
        \text{reshape}(\cos{\vw}, (d_{out}, d_{inp})).
 \end{equation}


\subsection{SLM Monarch layer}

The second free-space layer is inspired by digital Monarch matrices, which are structured matrices designed as parameter-efficient and expressive alternatives to dense linear layers~\cite{DaoMonarch,qiu2024compute}.
A Monarch matrix is defined by the product $PLP^TR$, where $P$ is a permutation matrix transforming from row-major to column-major ordering, and $L$ and $R$ are block-diagonal matrices given by: $L=\oplus_{\beta=1}L^{(\beta)}$, $R=\oplus_{\gamma=1}R^{(\gamma)}$, where $\beta$ and $\gamma$ mark the blocks within the block-diagonals.

The corresponding optical implementation, inspired by the block structure of Monarch matrices, is depicted in Figure~\ref{fig:physical_layers}C.
This implementation leverages multiple SLM-based MVM setups for block matrix multiplication.
However, our optical Monarch layer deviates from traditional digital Monarch matrices due to the complex-valued matrices $L$ and $R$ attainable with free-space optics.
Moreover, to facilitate multiplication by non-square matrices, the block matrices $L^{(\beta)}$ and $R^{(\gamma)}$ can be rectangular, having dimensions $n_L^{(out)}\times n_L^{(inp)}$ and $n_R^{(out)}\times n_R^{(inp)}$, respectively.
The block dimensions relate directly to the provided input  and output sizes $d_{inp}$ and $d_{out}$.
These dimensions are decomposed into products of two nearest powers of two: $d_{inp}=b_Rn_R^{(inp)}$, $d_{out}=b_Ln_L^{(out)}$, where $b_R$ and $b_L$ indicate the number of blocks within matrices $R$ and $L$, respectively. 
The remaining dimensions are determined by matching the output dimension of $R$ to the input dimension of $L$, resulting in $n_R^{(out)}=b_L$ and $n_L^{(inp)}=b_R$.

The resulting output of the optical Monarch layer is:
\begin{equation}
    y_{jl} = \text{Re} \left(
        \sum_{i=1}^{b_1} L^{(l)}_{ji}
        \sum_{k=1}^{n_1^{(inp)}} R^{(i)}_{lk}
    \right) \vx_{ik},
    \quad
        j = 1, \ldots, n_2^{(out)};
        \,
        l = 1, \ldots, b_2,
\end{equation}
with $R^{(i)}_{lk}=\frac{1}{\sqrt{n_1^{(inp)}}}e^{i\theta_{lk}^{(R)}}$ and $L^{(l)}_{ji}=\frac{1}{\sqrt{n_2^{(inp)}}}e^{i\theta_{ji}^{(L)}}$. As in the simpler SLM layer, the phase shifts $\theta_{lk}^{(R)}$ and $\theta_{ji}^{(L)}$ constitute the trainable weights $\vw$ of the optical Monarch layer.



\subsection{Planar programmable meshes}

Another our model of the optical layers describes the programmable interferometer circuits proposed in~\cite{ClementsDesign} and \cite{Hamerly_3MZI}. Such programmable circuits are  widely used in classical and quantum information processing~\cite{MZIquantum}. These circuits have the form of planar meshes consisting of $2\times2$ blocks $B(\theta,\varphi)$, each of which is independently programmed by tunable phase-shifts $\theta$ and $\varphi$. We consider two types of blocks, namely, the Mach-Zehnder interferometer (MZIs) and $3$-splitter MZI ($3$-MZI), depicted in  Figure~\ref{fig:physical_layers}D (i) and (ii), respectively.
Each MZI is parametrized by independent angles $\theta$ and $\varphi$ specified by the adjustable phase-shifts.
The MZI and $3$-MZI block transforms the incoming amplitudes according to the transfer matrices:
\begin{eqnarray*}
    \mB_{\text{MZI}}(\theta, \varphi) = e^{i\theta/2}\left(
        \begin{array}{cc}
           e^{i\varphi}\sin(\theta/2)  &  \cos(\theta/2)
           \\
           e^{i\varphi}\cos(\theta/2)  & -\sin(\theta/2)
        \end{array}
    \right),
\end{eqnarray*}
\begin{eqnarray*}
    \mB_{\text{3MZI}}(\theta, \varphi)=
    \frac{e^{i\frac{\varphi+\theta}{2}}}{\sqrt{2}}
    {\left(
    \begin{array}{cc}
        \!-\cos\frac{\theta-\varphi}{2}+i\sin\frac{\theta+\varphi}{2}\! & \!-\sin\frac{\theta-\varphi}{2}+i\cos\frac{\theta+\varphi}{2}\!  \\
        \!\sin\frac{\theta-\varphi}{2}+i\cos\frac{\theta+\varphi}{2}\!
         & 
         \!-\cos\frac{\theta-\varphi}{2}-i\sin\frac{\theta+\varphi}{2}\!
    \end{array}
    \right)\!.}\nonumber\\
\end{eqnarray*}

Accordingly, the $N \times N$ transfer matrix is obtained as a multiplication of $N$ MZI layers.
Together with the angles of the output phase-shifts $\Phi^{(out)}$, the number of all angles parameterizing the $2\times 2$ blocks sum up to $N^2$, exactly the number of  parameters to specify arbitrary $N \times N$ unitary matrices. 
These angle parameters are the trainable weights $\vw$.
To implement the MZI mesh layer with transfer matrices $\mAw^{(MZI)}$ of arbitrary dimensions, we embed them into an $N$-mode interferometer with $N = \text{max}(d_{out}, d_{inp})$ and make padding of the dangling inputs or outputs.
We note that the dependence of $A^{(MZI)}(\boldsymbol{w})$ on trainable parameters is very nonlinear and much more complicated than that of the MRR weight bank or SLM-based layers.

\section{Numerical experiments}
    \label{sec:experiments}

To demonstrate the practical utility of our hybrid training framework, \emph{astralora}, we conduct comprehensive experiments that integrate digital emulations of the photonic layers described in the previous section.
These software-based models accurately replicate the input-output behavior and, crucially, the black-box parameter-to-matrix mapping of their physical counterparts (e.g., MRRs, SLMs, MZI meshes). 
This approach allows for controlled, scalable validation of our framework without the experimental variability inherent to physical hardware.
We emphasize that while the simulations themselves are computationally intensive, an actual photonic implementation would perform these operations without such a digital bottleneck.
Our evaluation\footnote{
    We note that the digital emulation of the MZI mesh's transfer function (``mzi'' and ``3-mzi'' layers) is computationally expensive, rendering training prohibitively slow for our large-scale experiments.
    Therefore, we exclude them from the audio classification and text generation tasks, but present results for the simpler task of image classification.
} spans multiple domains: image classification, audio analysis, and text generation, confirming the framework's adaptability to real-world constraints and its ability to maintain competitive accuracy across diverse tasks.

\subsection{CIFAR-10 image classification}

 \begin{figure}[h!]
     \begin{center}
     \subfloat[\texttt{matvec}]{
         \includegraphics[width=.45\linewidth]{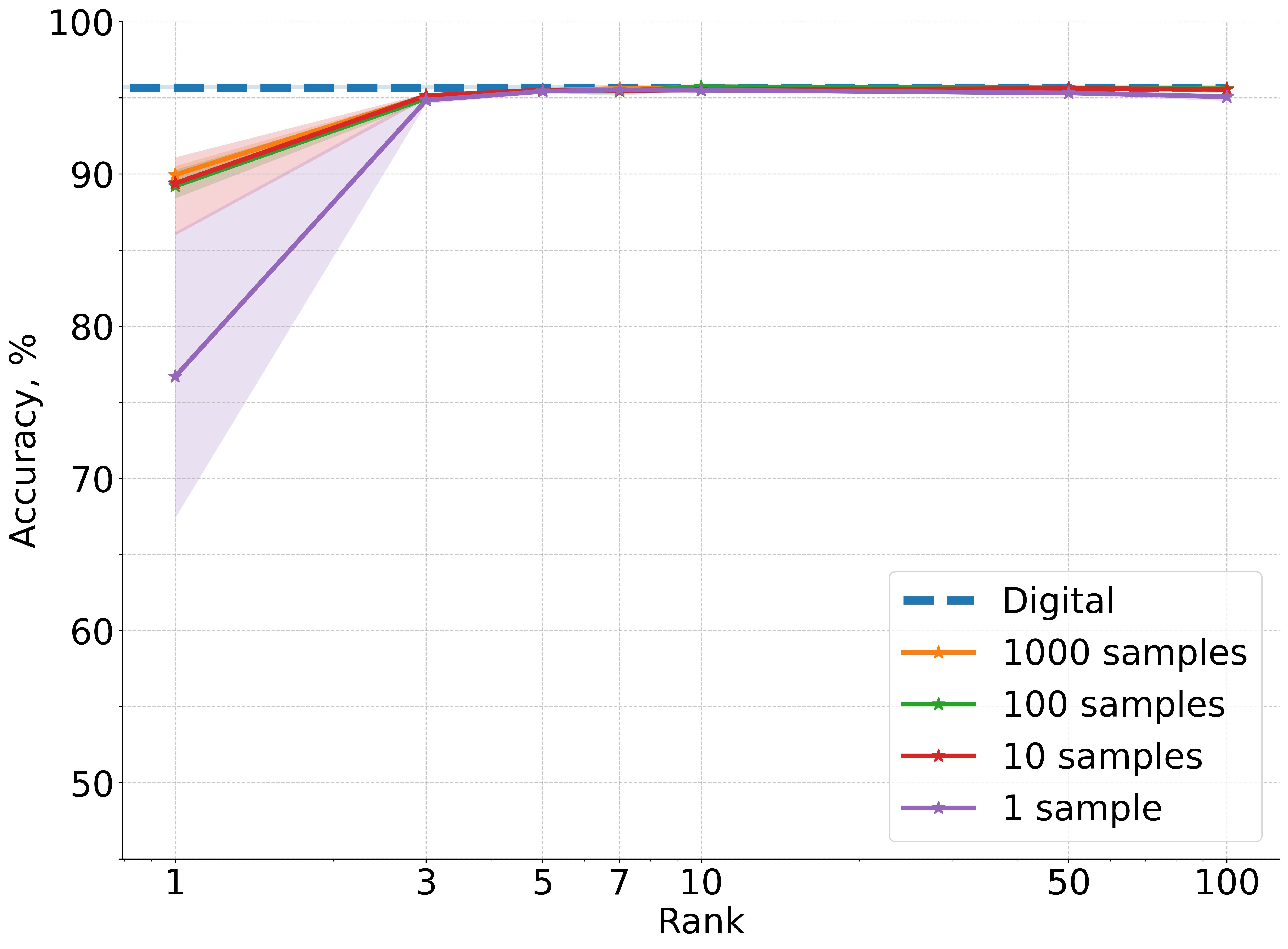}}
     \hfill
     \subfloat[\texttt{mrr}]{
         \includegraphics[width=.45\linewidth]{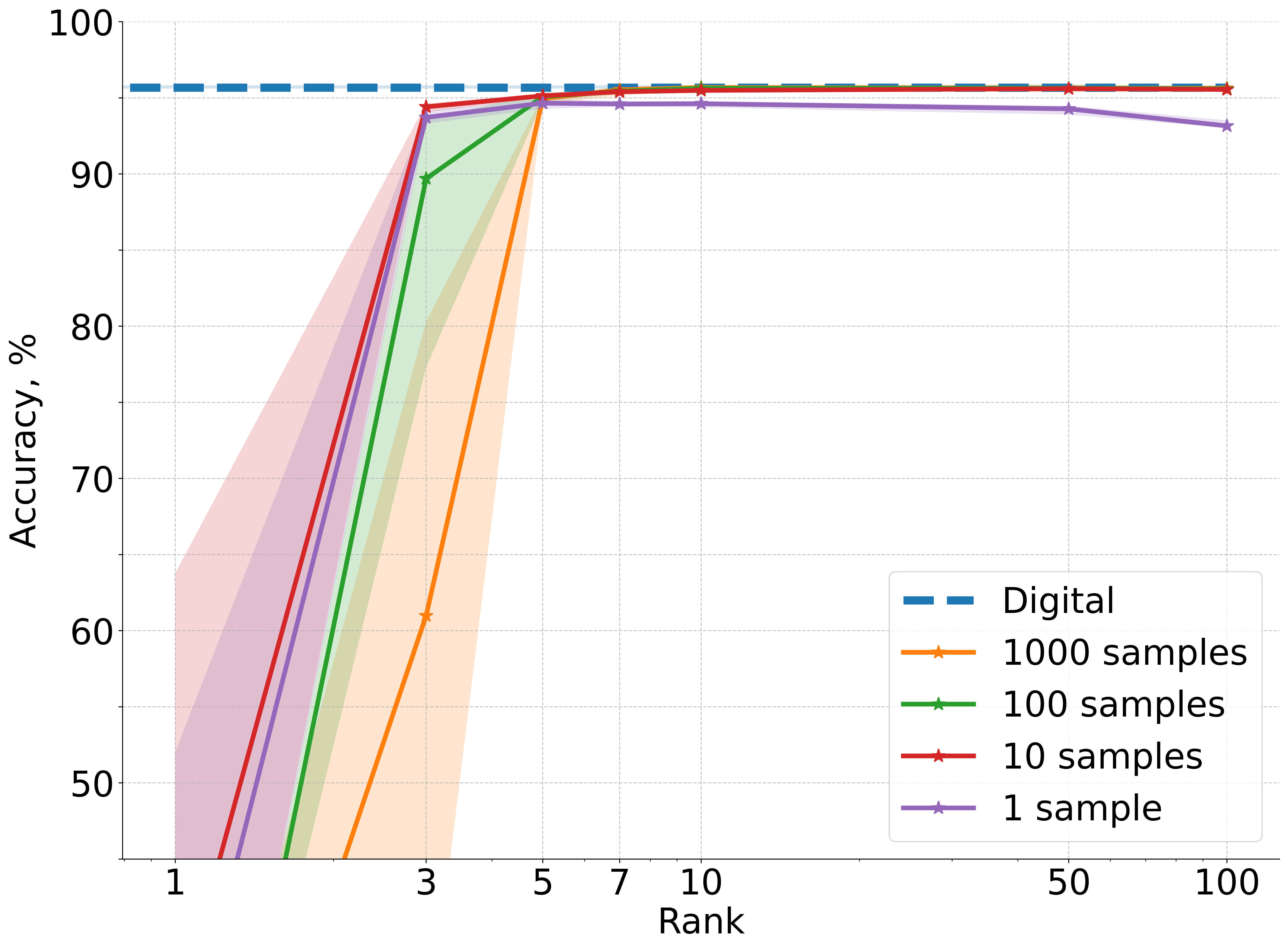}}
     \\
     \subfloat[\texttt{slm}]{
         \includegraphics[width=.45\linewidth]{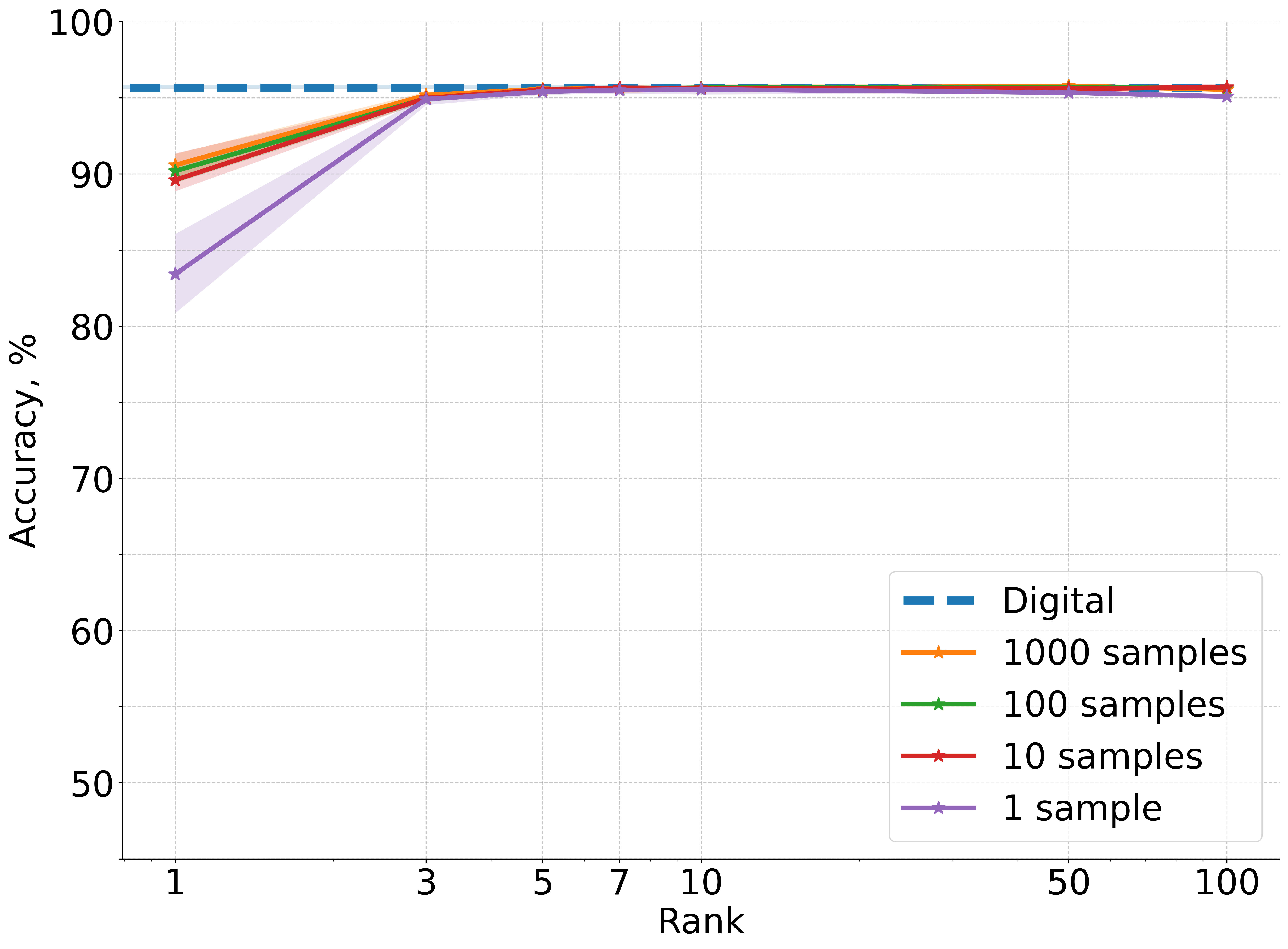}}
     \hfill
     \subfloat[\texttt{monarch}]{
         \includegraphics[width=.45\linewidth]{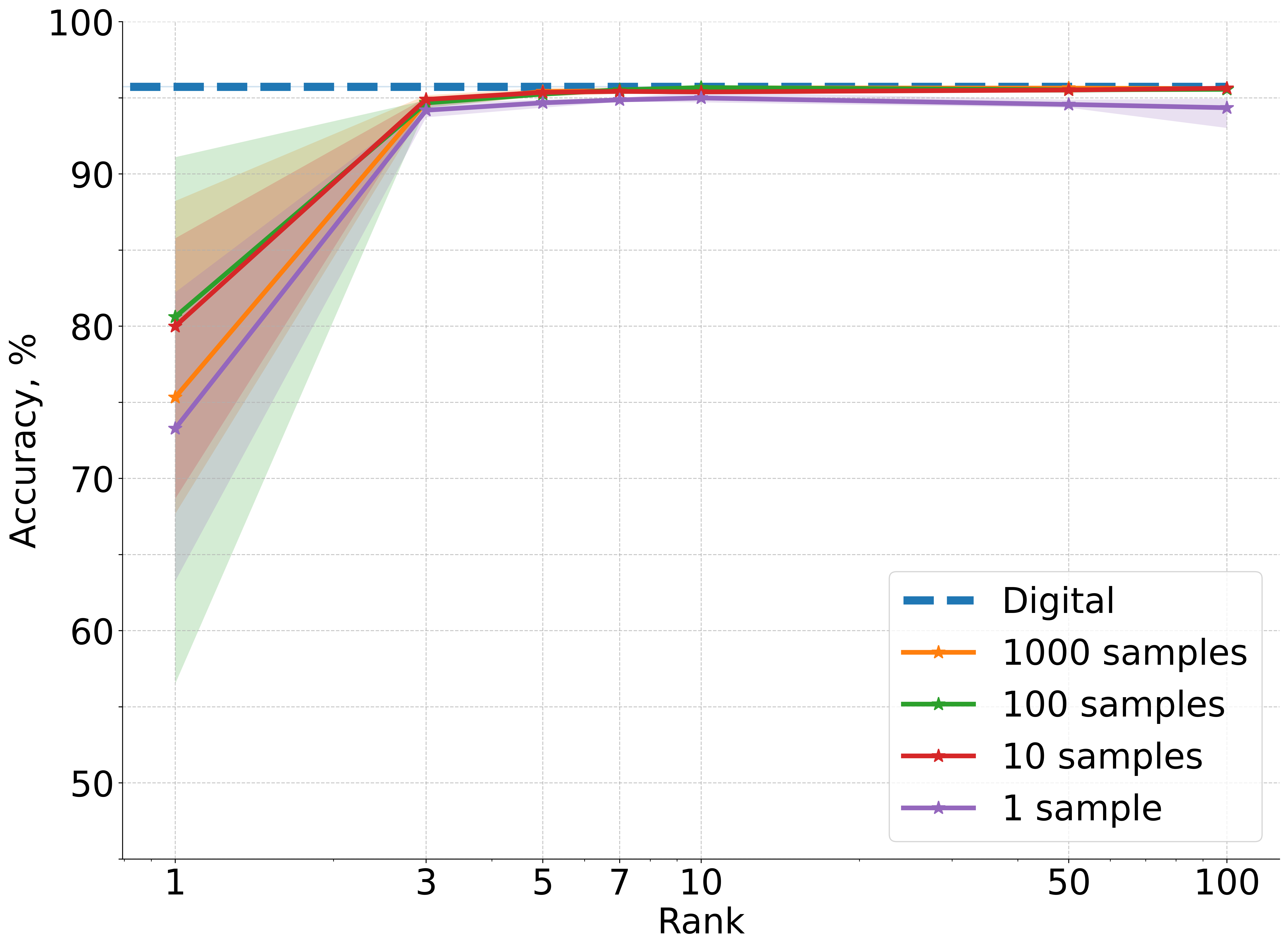}}
     \\
     \subfloat[\texttt{mzi}]{
         \includegraphics[width=.45\linewidth]{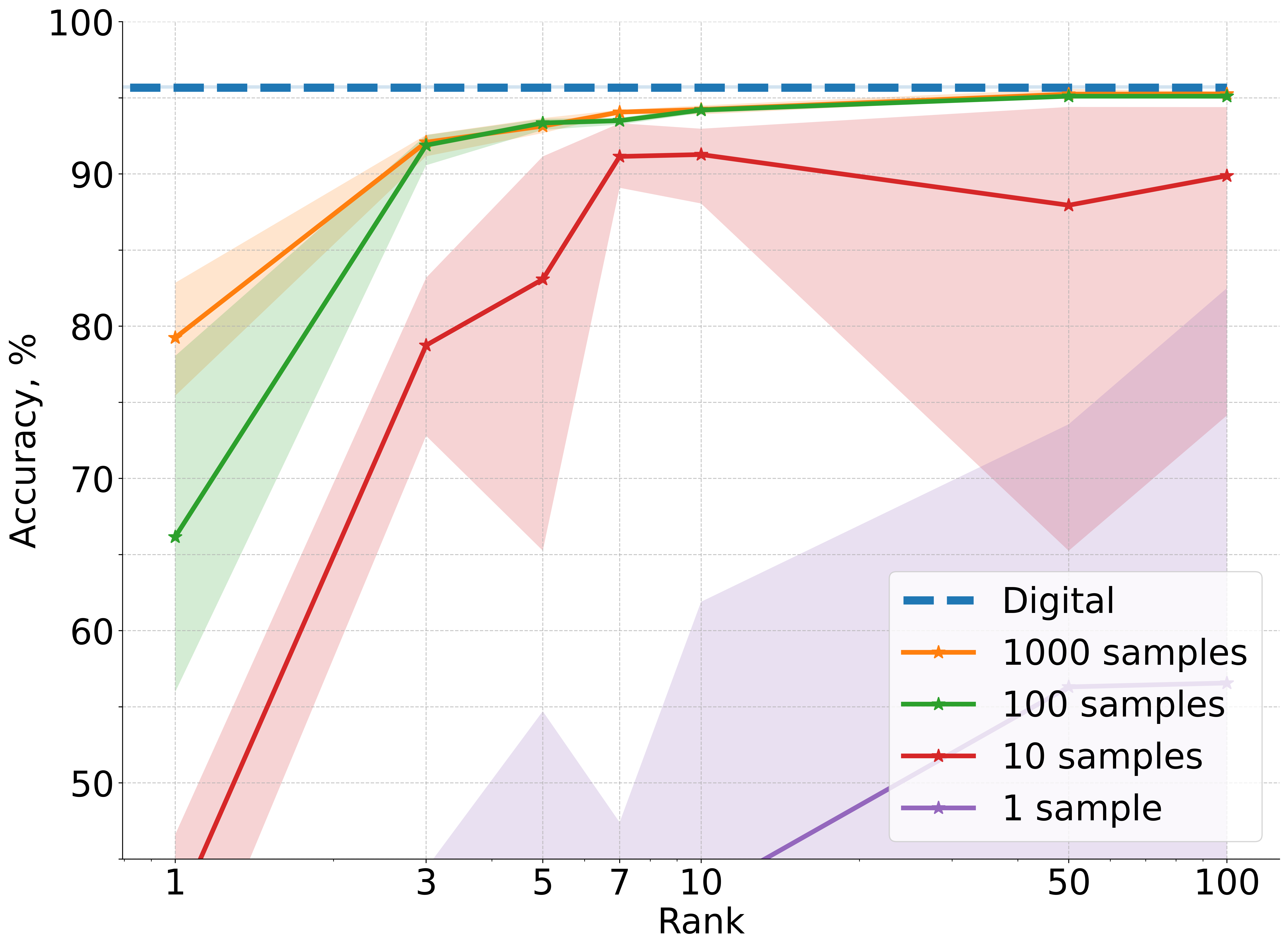}}
     \hfill
     \subfloat[\texttt{3-mzi}]{
        \includegraphics[width=.45\linewidth]{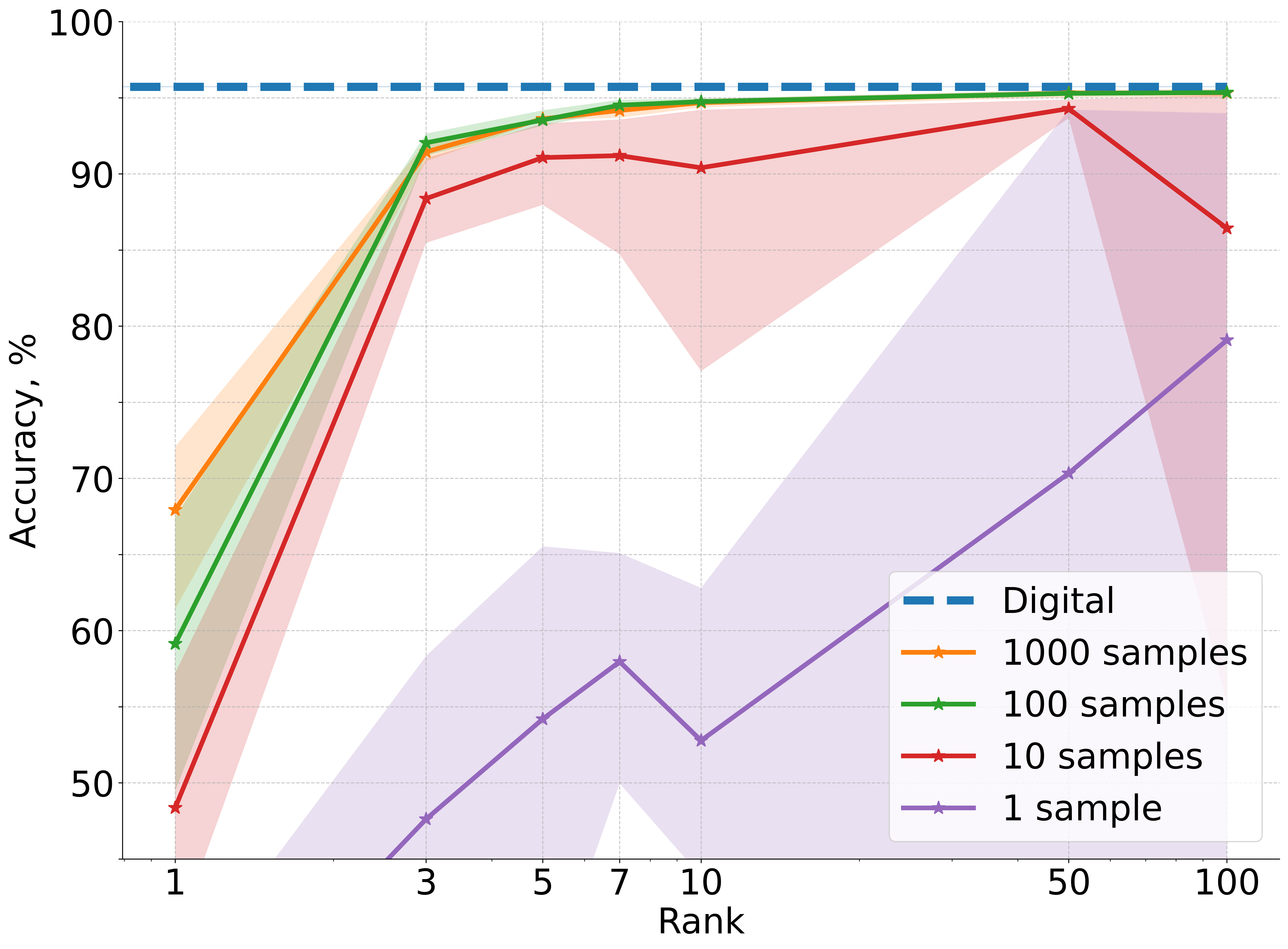}}
    \end{center}
    \caption{Accuracy results averaged over five independent runs for CIFAR-10 image classification using a deep convolutional neural network architecture where one linear layer is replaced with non-differentiable physical photonic layer.}
    \label{fig:results_airbench_cifar}
\end{figure}


We validate our framework on the well-known CIFAR-10 image classification benchmark~\cite{krizhevsky2009learning}, using a deep convolutional neural network.
\rewch{Such image classification tasks underpin a wide range of computer vision applications~\cite{vinay2023comparative, gao2025arctic, gao2025sfa, li2023sar}.} 
The base architecture\footnote{
    Our pipeline is based on \url{https://github.com/KellerJordan/cifar10-airbench}.
} consists of three convolutional blocks with residual connections, followed by global average pooling.
The final components are a classification block consisting of the primary linear layer ($512 \times 1024$) and an output linear layer ($1024 \times 10$).
We replace the primary linear layer with one of our BB layers (``matvec'', ``mrr'', ``slm'', ``monarch'', ``mzi'', and ``3-mzi'') and perform computations for various ranks $r$ of the SM model and varying numbers of allowed requests to the BB during the update of its parameters ($M_{bb}$) and during the update of the SM parameters $M_{sm}$ (for simplicity, we fix $M_{bb} = M_{sm} = M$).
For all combinations of $r=1, \, 3, \, 5, \, 7, \, 10, \, 50, \, 100$ and $M = 1, \, 10, \, 100, \, 1000$, we perform the computation five times for different random seeds and report the average result along with its boundary values in Figure~\ref{fig:results_airbench_cifar}.
For comparison, we also show the training result for the original model (referred to as the ``Digital'' curve) without replacing layers, trained with standard gradient-based optimization.
Results demonstrate that all photonic layer types achieve near-digital accuracy when sufficient rank ($r \geq 10$) and hardware queries ($M \geq 100$) are used, thus validating the effectiveness of the \emph{astralora} framework.

\subsection{Audio classification problem}

We further evaluate our framework on the audio classification problem using the UrbanSound8K dataset~\cite{salamon2014dataset} with the ECAPA-TDNN architecture~\cite{desplanques2020ecapa}\footnote{
    Our pipeline is based on \url{https://github.com/speechbrain/speechbrain}.
}.
This problem presents unique challenges due to the temporal nature of audio data and the need for precise feature extraction.
To simulate hardware integration, we replace a critical component in the ECAPA-TDNN feature extractor, i.e., the linear layer ($6144 \times 192$), with a BB layer while maintaining the original channel dimensions.
We fix $M_{bb} = M_{sm} = 1000$ and perform the training for different rank values $r = 1, 5, 10, 50$.
The results are averaged over five independent runs and reported in the Table~\ref{tbl:results_ecapa_urbansound8k}.
\rewch{We present the weighted F1-score as the primary metric, which accounts for class imbalance in the dataset and provides a more comprehensive evaluation than accuracy alone. 
The weighted F1-score is computed from the confusion matrix as the support-weighted average of per-class F1-scores.}

\begin{table}[t!]
\centering
\caption{
    \rewch{Weighted F1-score (\%)} for the audio classification problem using the UrbanSound8K dataset with the ECAPA-TDNN architecture.
    We consider different ranks of the SM and four BB layer types.
    All results are averaged over five independent runs.
    Reference accuracy of the digital model is $77.46\%$.
}
\label{tbl:results_ecapa_urbansound8k}
\begin{tabular}{l c c c c}
\toprule
Rank &
\multicolumn{4}{c}{Weighted F1-score (\%)}
\\
\cmidrule(lr){2-5}
of the SM          &
\texttt{matvec}    &
\texttt{mrr}       &
\texttt{slm}       &
\texttt{monarch}
\\
\midrule
1  & 70.82 & 59.10 & 69.30 & 56.63 \\
5  & 76.14 & 71.41 & 74.85 & 74.45 \\
10 & 74.93 & 74.00 & 78.47 & 74.13 \\
50 & 75.60 & 76.97 & 78.95 & 77.70 \\
\bottomrule
\end{tabular}
\end{table}

\begin{table}[t!]
\rewchinp
\centering
\caption{\rewch{Weighted F1-score (\%) for the audio classification problem using the UrbanSound8K dataset with the ECAPA-TDNN architecture. Calculations are presented for different computation modes at SM rank $r=10$. All results are averaged over five independent runs. The basic results (ZO+SM) match Table~\ref{tbl:results_ecapa_urbansound8k}.}}
\label{tbl:ablation_ecapa}
\begin{tabular}{ccccc}
\toprule
Approach
&
\texttt{matvec}
&
\texttt{mrr}
&
\texttt{slm}
&
\texttt{monarch}
\\ \midrule

ZO+SM & 74.93 & 74.00 & 78.47 & 74.13
\\ \hline
ZO-only & 76.15 & 74.95 & 77.64 & 78.41
\\ \hline
SM-only & 76.66 & 76.74 & 76.80 & 77.71
\\ \bottomrule
\end{tabular}
\end{table}

\begin{table}[t!]
\rewchinp
\centering
\caption{\rewch{Weighted F1-score (\%) for the audio classification problem using the UrbanSound8K dataset with the ECAPA-TDNN architecture. Calculations are presented for the ``slm'' layer under weight quantization. All results are averaged over five independent runs. Full-precision (32-bit) results match Table~\ref{tbl:results_ecapa_urbansound8k}.}}
\label{tbl:slm_quantization_results}
\begin{tabular}{cccccc}
\toprule
Rank & 2-bit & 3-bit & 4-bit & 8-bit & 32-bit
\\ \midrule

1  &  9.83 & 64.99 & 72.87 & 69.98 & 69.30 \\
5  & 18.47 & 73.98 & 72.67 & 75.28 & 74.85 \\
10 & 23.10 & 74.82 & 75.33 & 75.31 & 78.47 \\
50 & 37.66 & 76.52 & 76.47 & 75.60 & 78.95 \\

\bottomrule
\end{tabular}
\end{table}

The results, summarized in Table~\ref{tbl:results_ecapa_urbansound8k}, demonstrate the consistent effectiveness of the \emph{astralora} framework across diverse physical layer implementations (``matvec'', ``mrr'', ``slm'', and ``monarch'') for the audio classification task.
For SM rank $r = 50$, all configurations achieve weighted F1-scores within $2\%$ of the digital baseline, confirming effective end-to-end training despite the non-differentiability of the physical layers. 
Approximations with extremely low rank values of $r=1$ lead to noticeable degradation, due to limited expressiveness in capturing the BB's nonlinear parameter-to-matrix mapping.
However, performance recovers rapidly with increasing rank, with the ``slm'' and ``monarch'' layers nearing or matching the baseline at $r=50$.
This indicates that our dynamic low-rank SM, updated via the I-PSI algorithm, effectively approximates gradient flow through complex photonic operations, enabling convergence comparable to gradient-based methods.

To assess the contributions of \emph{astralora}'s core components, i.e., ZO optimization for BB parameter updates and surrogate modeling (SM) for input gradient estimation, we compare the full framework (ZO+SM) against variants using only ZO (replacing the SM with straight-through estimation during backpropagation) or only SM (replacing ZO with synthetic gradient updates for BB parameters).
Results at $r=10$ in the Table~\ref{tbl:ablation_ecapa} reveal that both individual components contribute meaningfully to performance, with the full ZO+SM framework achieves competitive results across all layer types, and is even better in the case of the ``slm'' layer.
These findings underscore the synergy between ZO and SM: ZO handles non-differentiable updates efficiently, while SM ensures stable upstream gradient flow.

\rewch{Additionally, to demonstrate robustness to hardware nonlinearities, such as limited precision in parameter control (e.g., due to discrete phase shifts in photonic devices) we evaluate performance under weight quantization of the BB parameters $\omega$.
We quantize $\omega$ to 2-bit, 3-bit, 4-bit, and 8-bit precision (simulating resolution constraints), comparing against the full-precision (32-bit) baseline, using the ``slm'' layer as a representative hardware-realistic case.
As shown in Table~\ref{tbl:slm_quantization_results}, even aggressive quantization (3-bit) maintains reasonable performance at higher ranks ($r \geq 10$), with degradation primarily at low ranks due to amplified approximation errors in the SM.
At 4-bit and above, scores are within 2--3\% of full precision.}

Overall, these experiments validate the \emph{astralora} framework and present its applicability to complex audio classification problem. 
\rewch{The framework's ability to achieve near-baseline weighted F1-scores, even with quantized parameters, highlights its potential for deploying hybrid models under realistic hardware nonlinearities and low-precision regimes.}

\subsection{Large-scale language modeling}

To rigorously validate the scalability and robustness of our framework, we conducted large-scale language modeling experiments using a GPT-2-like architecture (417M parameters)~\cite{radford2019language}\footnote{
    Our pipeline is based on \url{https://github.com/KellerJordan/modded-nanogpt}.
} trained on a subset of the FineWeb dataset~\cite{penedo2024the}.
This task presents a formidable challenge due to the model's size, the complexity of the natural language data, and the need for precise gradient approximations across multiple non-differentiable components.

We designed two distinct stress tests. In the first experiment (Table~\ref{tbl:results_nanogpt_fineweb}), we replaced only the first linear layer (of size $1536 \times 6144$) within one or more MLP blocks.
The blocks targeted for replacement were selected starting from the end of the network; thus, replacing one block means modifying the last MLP block, replacing four blocks modifies the last four, and so on.
This setup isolates the impact of introducing a black-box layer into the linear transformation part of the MLP.

In the second, more ambitious experiment (Table~\ref{tbl:results_nanogpt_fineweb_feedforward}), we replaced entire MLP blocks with black-box equivalents.
Here both linear layers and the intermediate nonlinearity within the block were treated as a single, monolithic black-box component.
Again, replacement proceeded from the end of the network. 
This scenario tests the framework's ability to handle a much more complex and deeply non-differentiable function.

\begin{table}[t!]
\centering
\caption{
    Validation perplexity for the language modeling with GPT-2--like architecture on FineWeb dataset with $1$, $4$, $8$, and $12$ linear layers in MLP blocks replaced by BB.
}
\label{tbl:results_nanogpt_fineweb}
\begin{tabular}{l l c c c c}
\toprule
\# of replaced           &
\# of digital            &
\multicolumn{4}{c}{Perplexity}
\\
\cmidrule(lr){3-6}
layers           &
parameters       &
\texttt{matvec}  &
\texttt{mrr}     &
\texttt{slm}     &
\texttt{monarch}
\\
\midrule
0  & $417M$ & \multicolumn{4}{c}{23.80}          \\
1  & $408M$ & $24.04$ & $24.28$ & $24.04$ & $24.28$ \\
4  & $379M$ & $25.02$ & $25.27$ & $25.02$ & $25.02$ \\
8  & $342M$ & $26.84$ & $27.11$ & $26.84$ & $27.11$ \\
12 & $304M$ & $30.26$ & $30.87$ & $30.26$ & $30.56$ \\
\bottomrule
\end{tabular}
\end{table}

\begin{table}[t!]
\centering
\caption{
    Validation perplexity for the language modeling with GPT-2--like architecture on FineWeb dataset with $1$, $4$, $8$, and $12$ MLP blocks replaced by BB.
}
\label{tbl:results_nanogpt_fineweb_feedforward}
\begin{tabular}{l l c c c c}
\toprule
\# of replaced           &
\# of digital            &
\multicolumn{4}{c}{Perplexity}
\\
\cmidrule(lr){3-6}
blocks           &
parameters       &
\texttt{matvec}  &
\texttt{mrr}     &
\texttt{slm}     &
\texttt{monarch}
\\
\midrule
0  & $417M$ & \multicolumn{4}{c}{23.80}          \\
1  & $398M$ & $24.28$ & $25.02$ & $24.28$ & $24.53$ \\
4  & $342M$ & $25.02$ & $25.53$ & $25.27$ & $25.02$ \\
8  & $266M$ & $27.11$ & $28.50$ & $27.11$ & $28.21$ \\
12 & $190M$ & $31.18$ & $31.50$ & $30.87$ & $32.13$ \\
\bottomrule
\end{tabular}
\end{table}

For both experiments, we fixed the query budgets to $M_{bb}=100$ for parameter updates and $M_{sm}=1000$ for surrogate model refinement.
The results, presented in Tables~\ref{tbl:results_nanogpt_fineweb} and~\ref{tbl:results_nanogpt_fineweb_feedforward}, demonstrate the consistent effectiveness of the \emph{astralora} framework.
In both setups, we observe a graceful degradation in performance as the number of replaced components increases.
This is expected, as each black-box layer introduces a small approximation error in the gradient.
The fact that the network remains trainable even when $12$ entire blocks are replaced (reducing the number of digitally updated parameters from $417M$ to $190M$) is a strong testament to the robustness of our approach.
We also note that the performance is remarkably consistent across all four physical layer types (``matvec'', ``mrr'', ``slm'', and ``monarch''), highlighting the general applicability of our method.


\section{Conclusions}
    \label{sec:conclusions}

In this work, we introduce \emph{astralora}, a general framework for training hybrid neural networks that incorporate non-differentiable physical components such as photonic layers.
By coupling stochastic zeroth-order optimization for hardware parameter updates with a dynamically refined low-rank surrogate model for gradient propagation, \emph{astralora} enables efficient end-to-end optimization without reliance on task-specific heuristics or restrictive architectural modifications. 
Our implicit projector-splitting integrator ensures lightweight surrogate updates with minimal hardware queries, effectively bridging the gap between physical device constraints and the demands of modern deep learning pipelines.

Extensive experiments across computer vision, speech recognition, and large-scale language modeling confirm that \emph{astralora} consistently achieves near-digital baseline accuracy, demonstrating its robustness across diverse modalities and physical implementations.
These results clearly highlight that accurate backpropagation is not strictly necessary for training hybrid networks with non-differentiable layers, provided efficient surrogate strategies are successfully employed.
Overall, \emph{astralora} provides a practical route for embedding physical layers into AI systems, thereby contributing toward the long-term goal of hardware-aware, energy-efficient, and high-performance machine learning.

\section*{Acknowledgements}

The work was supported by the grant for research centers in the field of AI provided by the Ministry of Economic Development of the Russian Federation in accordance with the agreement 000000C313925P4F0002 and the agreement with Skoltech №139-10-2025-033.

\bibliographystyle{elsarticle-num} 
\bibliography{biblio}

@article{ccarpinliouglu2025genetically,
  title={Genetically programmable optical random neural networks},
  author={{\c{C}}arp{\i}nl{\i}o{\u{g}}lu, Bora and Te{\u{g}}in, U{\u{g}}ur},
  journal={Communications Physics},
  volume={8},
  number={1},
  pages={349},
  year={2025},
  publisher={Nature Publishing Group UK London}
}

@article{chaubard2025scaling,
  title={Scaling Recurrent Neural Networks to a Billion Parameters with Zero-Order Optimization},
  author={Chaubard, Francois and Kochenderfer, Mykel},
  journal={arXiv preprint arXiv:2505.17852},
  year={2025}
}

@article{chen2024enhancing,
  title={Enhancing zeroth-order fine-tuning for language models with low-rank structures},
  author={Chen, Yiming and Zhang, Yuan and Cao, Liyuan and Yuan, Kun and Wen, Zaiwen},
  journal={arXiv preprint arXiv:2410.07698},
  year={2024}
}

@inproceedings{gooneratne2020low,
  title={Low-rank gradient approximation for memory-efficient on-device training of deep neural network},
  author={Gooneratne, Mary and Sim, Khe Chai and Zadrazil, Petr and Kabel, Andreas and Beaufays, Fran{\c{c}}oise and Motta, Giovanni},
  booktitle={ICASSP 2020-2020 IEEE International Conference on Acoustics, Speech and Signal Processing (ICASSP)},
  pages={3017--3021},
  year={2020},
  organization={IEEE}
}

@article{chen2023deepzero,
  title={Deepzero: Scaling up zeroth-order optimization for deep model training},
  author={Chen, Aochuan and Zhang, Yimeng and Jia, Jinghan and Diffenderfer, James and Liu, Jiancheng and Parasyris, Konstantinos and Zhang, Yihua and Zhang, Zheng and Kailkhura, Bhavya and Liu, Sijia},
  journal={arXiv preprint arXiv:2310.02025},
  year={2023}
}

@inproceedings{fournier2023can,
  title={Can forward gradient match backpropagation?},
  author={Fournier, Louis and Rivaud, St{\'e}phane and Belilovsky, Eugene and Eickenberg, Michael and Oyallon, Edouard},
  booktitle={International Conference on Machine Learning},
  pages={10249--10264},
  year={2023},
  organization={PMLR}
}

@article{krizhevsky2009learning,
  title={Learning multiple layers of features from tiny images},
  author={Krizhevsky, Alex and Hinton, Geoffrey and others},
  year={2009},
  publisher={Toronto, ON, Canada}
}

@article{liu2018zeroth,
  title={Zeroth-order stochastic variance reduction for nonconvex optimization},
  author={Liu, Sijia and Kailkhura, Bhavya and Chen, Pin-Yu and Ting, Paishun and Chang, Shiyu and Amini, Lisa},
  journal={Advances in neural information processing systems},
  volume={31},
  year={2018}
}

@article{lubich2014projector,
  title={A projector-splitting integrator for dynamical low-rank approximation},
  author={Lubich, Christian and Oseledets, Ivan V},
  journal={BIT Numerical Mathematics},
  volume={54},
  number={1},
  pages={171--188},
  year={2014},
  publisher={Springer}
}

@article{malladi2023fine,
  title={Fine-tuning language models with just forward passes},
  author={Malladi, Sadhika and Gao, Tianyu and Nichani, Eshaan and Damian, Alex and Lee, Jason D and Chen, Danqi and Arora, Sanjeev},
  journal={Advances in Neural Information Processing Systems},
  volume={36},
  pages={53038--53075},
  year={2023}
}

@article{momeni2025training,
  title={Training of physical neural networks},
  author={Momeni, Ali and Rahmani, Babak and Scellier, Benjamin and Wright, Logan G and McMahon, Peter L and Wanjura, Clara C and Li, Yuhang and Skalli, Anas and Berloff, Natalia G and Onodera, Tatsuhiro and others},
  journal={Nature},
  volume={645},
  number={8079},
  pages={53--61},
  year={2025},
  publisher={Nature Publishing Group UK London}
}

@inproceedings{olaleke2021dynamic,
  title={Dynamic modeling of user preferences for stable recommendations},
  author={Olaleke, Oluwafemi and Oseledets, Ivan and Frolov, Evgeny},
  booktitle={Proceedings of the 29th ACM Conference on User Modeling, Adaptation and Personalization},
  pages={262--266},
  year={2021}
}

@article{qiu2024compute,
  title={Compute better spent: Replacing dense layers with structured matrices},
  author={Qiu, Shikai and Potapczynski, Andres and Finzi, Marc and Goldblum, Micah and Wilson, Andrew Gordon},
  journal={arXiv preprint arXiv:2406.06248},
  year={2024}
}

@article{refael2024adarankgrad,
  title={AdaRankGrad: Adaptive Gradient-Rank and Moments for Memory-Efficient LLMs Training and Fine-Tuning},
  author={Refael, Yehonathan and Svirsky, Jonathan and Shustin, Boris and Huleihel, Wasim and Lindenbaum, Ofir},
  journal={arXiv preprint arXiv:2410.17881},
  year={2024}
}

@article{refael2025sumo,
  title={{SUMO}: Subspace-Aware Moment-Orthogonalization for Accelerating Memory-Efficient {LLM} Training},
  author={Refael, Yehonathan and Smorodinsky, Guy and Tirer, Tom and Lindenbaum, Ofir},
  journal={arXiv preprint arXiv:2505.24749},
  year={2025}
}

@article{wang2024lora,
  title={Lora-ga: Low-rank adaptation with gradient approximation},
  author={Wang, Shaowen and Yu, Linxi and Li, Jian},
  journal={Advances in Neural Information Processing Systems},
  volume={37},
  pages={54905--54931},
  year={2024}
}

@article{zhao2024galore,
  title={Galore: Memory-efficient llm training by gradient low-rank projection},
  author={Zhao, Jiawei and Zhang, Zhenyu and Chen, Beidi and Wang, Zhangyang and Anandkumar, Anima and Tian, Yuandong},
  journal={arXiv preprint arXiv:2403.03507},
  year={2024}
}

@article{radford2019language,
  title={Language models are unsupervised multitask learners},
  author={Radford, Alec and Wu, Jeffrey and Child, Rewon and Luan, David and Amodei, Dario and Sutskever, Ilya and others},
  journal={OpenAI blog},
  volume={1},
  number={8},
  pages={9},
  year={2019}
}

@inproceedings{
  penedo2024the,
  title={The FineWeb Datasets: Decanting the Web for the Finest Text Data at Scale},
  author={Guilherme Penedo and Hynek Kydl{\'\i}{\v{c}}ek and Loubna Ben allal and Anton Lozhkov and Margaret Mitchell and Colin Raffel and Leandro Von Werra and Thomas Wolf},
  booktitle={The Thirty-eight Conference on Neural Information Processing Systems Datasets and Benchmarks Track},
  year={2024},
  urlold={https://openreview.net/forum?id=n6SCkn2QaG}
}

@article{desplanques2020ecapa,
  title={{ECAPA-TDNN}: Emphasized channel attention, propagation and aggregation in tdnn based speaker verification},
  author={Desplanques, Brecht and Thienpondt, Jenthe and Demuynck, Kris},
  journal={arXiv preprint arXiv:2005.07143},
  year={2020}
}

@inproceedings{salamon2014dataset,
  title={A dataset and taxonomy for urban sound research},
  author={Salamon, Justin and Jacoby, Christopher and Bello, Juan Pablo},
  booktitle={Proceedings of the 22nd ACM international conference on Multimedia},
  pages={1041--1044},
  year={2014}
}

@Article{OnChipDiffractive,
    author={Fu, Tingzhao
    and Zang, Yubin
    and Huang, Yuyao
    and Du, Zhenmin
    and Huang, Honghao
    and Hu, Chengyang
    and Chen, Minghua
    and Yang, Sigang
    and Chen, Hongwei},
    title={Photonic machine learning with on-chip diffractive optics},
    journal={Nature Communications},
    year={2023},
    month={Jan},
    day={05},
    volume={14},
    number={1},
    pages={70},
    issn={2041-1723},
    doiold={10.1038/s41467-022-35772-7},
    urlold={https://doi.org/10.1038/s41467-022-35772-7}
}

@inproceedings{L2ight,
     author = {Gu, Jiaqi and Zhu, Hanqing and Feng, Chenghao and Jiang, Zixuan and Chen, Ray and Pan, David},
     booktitle = {Advances in Neural Information Processing Systems},
     editor = {M. Ranzato and A. Beygelzimer and Y. Dauphin and P.S. Liang and J. Wortman Vaughan},
     pages = {8649--8661},
     publisher = {Curran Associates, Inc.},
     title = {L2ight: Enabling On-Chip Learning for Optical Neural Networks via Efficient in-situ Subspace Optimization},
     urlold = {https://proceedings.neurips.cc/paper_files/paper/2021/file/48aedb8880cab8c45637abc7493ecddd-Paper.pdf},
     volume = {34},
     year = {2021}
}

@misc{TensorCompressedTraining,
      title={Tensor-Compressed Back-Propagation-Free Training for (Physics-Informed) Neural Networks}, 
      author={Yequan Zhao and Xinling Yu and Zhixiong Chen and Ziyue Liu and Sijia Liu and Zheng Zhang},
      year={2023},
      eprintold={2308.09858},
      archivePrefix={arXiv},
      primaryClass={cs.LG},
      urlold={https://arxiv.org/abs/2308.09858}, 
}

@inproceedings{pMetaOnDevice, 
    series={KDD ’22},
    title={p-Meta: Towards On-device Deep Model Adaptation},
    urlold={http://dx.doi.org/10.1145/3534678.3539293},
    doiold={10.1145/3534678.3539293},
    booktitle={Proceedings of the 28th ACM SIGKDD Conference on Knowledge Discovery and Data Mining},
    publisher={ACM},
    author={Qu, Zhongnan and Zhou, Zimu and Tong, Yongxin and Thiele, Lothar},
    year={2022},
    month=aug, pages={1441–1451},
    collection={KDD ’22} 
}

@article{SimPhony,
  title={SimPhony: A Device-Circuit-Architecture Cross-Layer Modeling and Simulation Framework for Heterogeneous Electronic-Photonic AI System},
  author={Yin, Ziang and Zhang, Meng and Begovic, Amir and Huang, Rena and Zhang, Jeff and Gu, Jiaqi},
  journal={arXiv preprint arXiv:2411.13715},
  year={2024}
}

@article{insitu_backprop_integrated,
    author = {Sunil Pai  and Zhanghao Sun  and Tyler W. Hughes  and Taewon Park  and Ben Bartlett  and Ian A. D. Williamson  and Momchil Minkov  and Maziyar Milanizadeh  and Nathnael Abebe  and Francesco Morichetti  and Andrea Melloni  and Shanhui Fan  and Olav Solgaard  and David A. B. Miller },
    title = {Experimentally realized in situ backpropagation for deep learning in photonic neural networks},
    journal = {Science},
    volume = {380},
    number = {6643},
    pages = {398-404},
    year = {2023},
    doiold = {10.1126/science.ade8450},
    urlold = {https://www.science.org/doi/abs/10.1126/science.ade8450},
    eprintold= {https://www.science.org/doi/pdf/10.1126/science.ade8450},
}

@misc{dfa_optical_transformer,
      title={Streamlined optical training of large-scale modern deep learning architectures with direct feedback alignment}, 
      author={Ziao Wang and Kilian Müller and Matthew Filipovich and Julien Launay and Ruben Ohana and Gustave Pariente and Safa Mokaadi and Charles Brossollet and Fabien Moreau and Alessandro Cappelli and Iacopo Poli and Igor Carron and Laurent Daudet and Florent Krzakala and Sylvain Gigan},
      year={2025},
      eprintold={2409.12965},
      archivePrefix={arXiv},
      primaryClass={cs.ET},
      urlold={https://arxiv.org/abs/2409.12965}, 
}

@article{ffa_hinton,
  title={The forward-forward algorithm: Some preliminary investigations},
  author={Hinton, Geoffrey},
  journal={arXiv preprint arXiv:2212.13345},
  volume={2},
  number={3},
  pages={5},
  year={2022}
}

@article{FFA_optical,
    author = {Ilker Oguz and Junjie Ke and Qifei Weng and Feng Yang and Mustafa Yildirim and Niyazi Ulas Dinc and Jih-Liang Hsieh and Christophe Moser and Demetri Psaltis},
    journal = {Opt. Lett.},
    number = {20},
    pages = {5249--5252},
    publisher = {Optica Publishing Group},
    title = {Forward--forward training of an optical neural network},
    volume = {48},
    month = {Oct},
    year = {2023},
    urlold = {https://opg.optica.org/ol/abstract.cfm?URI=ol-48-20-5249},
    doiold = {10.1364/OL.496884},
}

@article{DNN_backprop_proposal,
    author = {Tiankuang Zhou and Lu Fang and Tao Yan and Jiamin Wu and Yipeng Li and Jingtao Fan and Huaqiang Wu and Xing Lin and Qionghai Dai},
    journal = {Photon. Res.},
    number = {6},
    pages = {940--953},
    publisher = {Optica Publishing Group},
    title = {In situ optical backpropagation training of diffractive optical neural networks},
    volume = {8},
    month = {Jun},
    year = {2020},
    urlold = {https://opg.optica.org/prj/abstract.cfm?URI=prj-8-6-940},
    doiold = {10.1364/PRJ.389553},
}

@Article{Hamerly_3MZI,
    author={Hamerly, Ryan
    and Bandyopadhyay, Saumil
    and Englund, Dirk},
    title={Asymptotically fault-tolerant programmable photonics},
    journal={Nature Communications},
    year={2022},
    tmpmonth={Nov},
    tmpday={29},
    volume={13},
    tmpnumber={1},
    tmppages={6831},
    issn={2041-1723},
    doiold={10.1038/s41467-022-34308-3},
    urlold={https://doi.org/10.1038/s41467-022-34308-3}
}

@article{onchip_forwardonly_englund,
    author={Bandyopadhyay, Saumil
    and Sludds, Alexander
    and Krastanov, Stefan
    and Hamerly, Ryan
    and Harris, Nicholas
    and Bunandar, Darius
    and Streshinsky, Matthew
    and Hochberg, Michael
    and Englund, Dirk},
    title={Single-chip photonic deep neural network with forward-only training},
    journal={Nature Photonics},
    year={2024},
    month={Dec},
    day={01},
    volume={18},
    number={12},
    pages={1335-1343},
    issn={1749-4893},
    doiold={10.1038/s41566-024-01567-z},
    urlold={https://doi.org/10.1038/s41566-024-01567-z}
}

@article{multiplexed_gd_proposal,
    author = {McCaughan, Adam N. and Oripov, Bakhrom G. and Ganesh, Natesh and Nam, Sae Woo and Dienstfrey, Andrew and Buckley, Sonia M.},
    title = {Multiplexed gradient descent: Fast online training of modern datasets on hardware neural networks without backpropagation},
    journal = {APL Machine Learning},
    volume = {1},
    number = {2},
    pages = {026118},
    year = {2023},
    month = {06},
    issn = {2770-9019},
    doiold = {10.1063/5.0157645},
    urlold = {https://doi.org/10.1063/5.0157645},
    eprint0 = {https://pubs.aip.org/aip/aml/article-pdf/doi/10.1063/5.0157645/18017061/026118\_1\_5.0157645.pdf},
}

@ARTICLE{MRR_weightbanks,
  author={Tait, Alexander N. and Wu, Allie X. and de Lima, Thomas Ferreira and Zhou, Ellen and Shastri, Bhavin J. and Nahmias, Mitchell A. and Prucnal, Paul R.},
  journal={IEEE Journal of Selected Topics in Quantum Electronics}, 
  title={Microring Weight Banks}, 
  year={2016},
  volume={22},
  number={6},
  pages={312-325},
  doiold={10.1109/JSTQE.2016.2573583}
}

@ARTICLE{CoherentXbar, 
    author={Giamougiannis, George and Tsakyridis, Apostolos and Ma, Yangjin and Totović, Angelina and Moralis-Pegios, Miltiadis and Lazovsky, David and Pleros, Nikos}, 
    journal={Journal of Lightwave Technology}, title={A Coherent Photonic Crossbar for Scalable Universal Linear Optics}, 
    year={2023}, 
    volume={41}, 
    number={8}, 
    pages={2425-2442}, 
    doiold={10.1109/JLT.2023.3234689}
}

@Article{MZIquantum,
    author={Dong, Mark
    and Zimmermann, Matthew
    and Heim, David
    and Choi, Hyeongrak
    and Clark, Genevieve
    and Leenheer, Andrew J.
    and Palm, Kevin J.
    and Witte, Alex
    and Dominguez, Daniel
    and Gilbert, Gerald
    and Eichenfield, Matt
    and Englund, Dirk},
    title={Programmable photonic integrated meshes for modular generation of optical entanglement links},
    journal={npj Quantum Information},
    year={2023},
    tmpmonth={Apr},
    day={27},
    volume={9},
    tmpnumber={1},
    tmppages={42},
    issn={2056-6387},
    doiold={10.1038/s41534-023-00708-6},
    urlold={https://doi.org/10.1038/s41534-023-00708-6}
}

@article{ClementsDesign,
    author = {William R. Clements and Peter C. Humphreys and Benjamin J. Metcalf and W. Steven Kolthammer and Ian A. Walmsley},
    journal = {Optica},
    number = {12},
    pages = {1460--1465},
    publisher = {Optica Publishing Group},
    title = {Optimal design for universal multiport interferometers},
    volume = {3},
    month = {Dec},
    year = {2016},
    urlold = {https://opg.optica.org/optica/abstract.cfm?URI=optica-3-12-1460},
    doiold = {10.1364/OPTICA.3.001460},
}

@article{TamuraSLM,
    author = {N. Tamura and J. C. Wyant},
    journal = {Opt. Eng.},
    number = {198},
    title = {Two-Dimensional Matrix Multiplication using Coherent Optical Techniques},
    volume = {18},
    year = {1979},
}

@article{LvovskySLM,
    author = {James Spall and Xianxin Guo and Thomas D. Barrett and A. I. Lvovsky},
    journal = {Opt. Lett.},
    number = {20},
    pages = {5752--5755},
    publisher = {Optica Publishing Group},
    title = {Fully reconfigurable coherent optical vector--matrix multiplication},
    volume = {45},
    month = {Oct},
    year = {2020},
    urlold = {https://opg.optica.org/ol/abstract.cfm?URI=ol-45-20-5752},
    doiold = {10.1364/OL.401675},
}

@misc{DaoMonarch,
      title={Monarch: Expressive Structured Matrices for Efficient and Accurate Training}, 
      author={Tri Dao and Beidi Chen and Nimit Sohoni and Arjun Desai and Michael Poli and Jessica Grogan and Alexander Liu and Aniruddh Rao and Atri Rudra and Christopher Ré},
      year={2022},
      eprintold={2204.00595},
      archivePrefix={arXiv},
      primaryClass={cs.LG},
      urlold={https://arxiv.org/abs/2204.00595}, 
}

@article{bente2025potential,
  title={The potential of multidimensional photonic computing},
  author={Bente, Ivonne and Taheriniya, Shabnam and Lenzini, Francesco and Br{\"u}ckerhoff-Pl{\"u}ckelmann, Frank and Kues, Michael and Bhaskaran, Harish and Wright, C David and Pernice, Wolfram},
  journal={Nature Reviews Physics},
  pages={1--12},
  year={2025},
  publisher={Nature Publishing Group UK London}
}

@article{sarker2021deep,
  title={Deep learning: a comprehensive overview on techniques, taxonomy, applications and research directions},
  author={Sarker, Iqbal H},
  journal={SN computer science},
  volume={2},
  number={6},
  pages={1--20},
  year={2021},
  publisher={Springer}
}

@article{moralis2022neuromorphic,
  title={Neuromorphic silicon photonics and hardware-aware deep learning for high-speed inference},
  author={Moralis-Pegios, Miltiadis and Mourgias-Alexandris, George and Tsakyridis, Apostolos and Giamougiannis, George and Totovic, Angelina and Dabos, George and Passalis, Nikolaos and Kirtas, Manos and Rutirawut, T and Gardes, FY and others},
  journal={Journal of Lightwave Technology},
  volume={40},
  number={10},
  pages={3243--3254},
  year={2022},
  publisher={IEEE}
}

@article{ahmed2025universal,
  title={Universal photonic artificial intelligence acceleration},
  author={Ahmed, Sufi R and Baghdadi, Reza and Bernadskiy, Mikhail and Bowman, Nate and Braid, Ryan and Carr, Jim and Chen, Chen and Ciccarella, Pietro and Cole, Matthew and Cooke, John and others},
  journal={Nature},
  volume={640},
  number={8058},
  pages={368--374},
  year={2025},
  publisher={Nature Publishing Group UK London}
}

@article{ma2025quantum,
  title={Quantum-limited stochastic optical neural networks operating at a few quanta per activation},
  author={Ma, Shi-Yuan and Wang, Tianyu and Laydevant, J{\'e}r{\'e}mie and Wright, Logan G and McMahon, Peter L},
  journal={Nature Communications},
  volume={16},
  number={1},
  pages={359},
  year={2025},
  publisher={Nature Publishing Group UK London}
}

@article{liao2023integrated,
  title={Integrated photonic neural networks: Opportunities and challenges},
  author={Liao, Kun and Dai, Tianxiang and Yan, Qiuchen and Hu, Xiaoyong and Gong, Qihuang},
  journal={ACS Photonics},
  volume={10},
  number={7},
  pages={2001--2010},
  year={2023},
  publisher={ACS Publications}
}

@article{montes2024fundamentals,
  title={Fundamentals and recent developments of free-space optical neural networks},
  author={Montes McNeil, Alexander and Li, Yuxiao and Zhang, Allen and Moebius, Michael and Liu, Yongmin},
  journal={Journal of Applied Physics},
  volume={136},
  number={3},
  year={2024},
  publisher={AIP Publishing}
}

@article{shastri2021photonics,
  title={Photonics for artificial intelligence and neuromorphic computing},
  author={Shastri, Bhavin J and Tait, Alexander N and Ferreira de Lima, Thomas and Pernice, Wolfram HP and Bhaskaran, Harish and Wright, C David and Prucnal, Paul R},
  journal={Nature Photonics},
  volume={15},
  number={2},
  pages={102--114},
  year={2021},
  publisher={Nature Publishing Group UK London}
}

@article{wang2025photonics,
  title={Photonics breakthroughs 2024: Nonlinear photonic computing at scale},
  author={Wang, Hao and Hu, Jianqi and Morandi, Andrea and Nardi, Alfonso and Xia, Fei and Li, Xuanchen and Savo, Romolo and Liu, Qiang and Grange, Rachel and Gigan, Sylvain},
  journal={IEEE Photonics Journal},
  year={2025},
  publisher={IEEE}
}

@article{yildirim2024nonlinear,
  title={Nonlinear processing with linear optics},
  author={Yildirim, Mustafa and Dinc, Niyazi Ulas and Oguz, Ilker and Psaltis, Demetri and Moser, Christophe},
  journal={Nature Photonics},
  volume={18},
  number={10},
  pages={1076--1082},
  year={2024},
  publisher={Nature Publishing Group UK London}
}

@article{moralis2024perfect,
  title={Perfect linear optics using silicon photonics},
  author={Moralis-Pegios, Miltiadis and Giamougiannis, George and Tsakyridis, Apostolos and Lazovsky, David and Pleros, Nikos},
  journal={Nature Communications},
  volume={15},
  number={1},
  pages={5468},
  year={2024},
  publisher={Nature Publishing Group UK London}
}

@article{najjar2024deep,
  title={Deep photonic network platform enabling arbitrary and broadband optical functionality},
  author={Najjar Amiri, Ali and Vit, Aycan Deniz and Gorgulu, Kazim and Magden, Emir Salih},
  journal={Nature Communications},
  volume={15},
  number={1},
  pages={1432},
  year={2024},
  publisher={Nature Publishing Group UK London}
}

@article{pai2019parallel,
  title={Parallel fault-tolerant programming of an arbitrary feedforward photonic network},
  author={Pai, Sunil and Williamson, Ian AD and Hughes, Tyler W and Minkov, Momchil and Solgaard, Olav and Fan, Shanhui and Miller, David AB},
  journal={arXiv preprint arXiv:1909.06179},
  year={2019}
}

@article{laporte2019highly,
  title={Highly parallel simulation and optimization of photonic circuits in time and frequency domain based on the deep-learning framework pytorch},
  author={Laporte, Floris and Dambre, Joni and Bienstman, Peter},
  journal={Scientific reports},
  volume={9},
  number={1},
  pages={5918},
  year={2019},
  publisher={Nature Publishing Group UK London}
}

@inproceedings{jiaqigu2021L2ight,
    title     = {L2ight: Enabling On-Chip Learning for Optical Neural Networks via Efficient in-situ Subspace Optimization},
    author    = {Jiaqi Gu and Hanqing Zhu and Chenghao Feng and Zixuan Jiang and Ray T. Chen and David Z. Pan},
    booktitle = {Conference on Neural Information Processing Systems (NeurIPS)},
    year      = {2021}
}

@article{zheng2023dual,
title={Dual adaptive training of photonic neural networks},
author={Zheng, Ziyang and Duan, Zhengyang and Chen, Hang and Yang, Rui and Gao, Sheng and Zhang, Haiou and Xiong, Hongkai and Lin, Xing},
journal={Nature Machine Intelligence},
pages={1--11},
year={2023},
publisher={Nature Publishing Group UK London}
}

@article{gao2025arctic,
  title={Arctic sea ice motion retrieval from multisource {SAR} images using a keypoint-free feature tracking algorithm},
  author={Gao, Tian and Lan, Chaozhen and Zhou, Chunxia and Zhang, Yongxian and Huang, Wenjun and Wang, Yiqiao and Wang, Longhao},
  journal={ISPRS Journal of Photogrammetry and Remote Sensing},
  volume={230},
  pages={258--274},
  year={2025},
  publisher={Elsevier}
}

@article{gao2025sfa,
  title={{SFA-Net}: A {SAM}-guided focused attention network for multimodal remote sensing image matching},
  author={Gao, Tian and Lan, Chaozhen and Huang, Wenjun and Wang, Sheng},
  journal={ISPRS Journal of Photogrammetry and Remote Sensing},
  volume={223},
  pages={188--206},
  year={2025},
  publisher={Elsevier}
}

@article{li2023sar,
  title={{SAR}--Optical Image Matching With Semantic Position Probability Distribution},
  author={Li, Liangzhi and Han, Ling and Liu, Ming and Gao, Kyle and He, Hongjie and Wang, Lanying and Li, Jonathan},
  journal={IEEE Transactions on Geoscience and Remote Sensing},
  volume={61},
  pages={1--15},
  year={2023},
  publisher={IEEE}
}

@article{gai2025interpretable,
  title={Interpretable unsupervised neural network structure for data clustering via differentiable reconstruction of {ONMF} and sparse autoencoder},
  author={Gai, Yongwei and Liu, Jinglei},
  journal={Neural Networks},
  pages={107504},
  year={2025},
  publisher={Elsevier}
}

@article{wang2025dual,
  title={Dual-space topological isomorphism and maximization of predictive diversity for unsupervised domain adaptation},
  author={Wang, Mengru and Liu, Jinglei},
  journal={IEEE Transactions on Image Processing},
  year={2025},
  publisher={IEEE}
}

@article{vinay2023comparative,
  title={A comparative study of convolutional neural networks and cybernetic approaches on {CIFAR-10} dataset},
  author={Vinay, SB and Balasubramanian, S},
  journal={International Journal of Machine Learning and Cybernetics (IJMLC)},
  volume={1},
  number={1},
  pages={1--13},
  year={2023}
}
\end{document}